\documentclass{article}

\usepackage[preprint]{neurips_2025}
\usepackage[utf8]{inputenc} % allow utf-8 input
\usepackage[T1]{fontenc}    % use 8-bit T1 fonts
\usepackage{hyperref}       % hyperlinks
\usepackage{url}            % simple URL typesetting (neurips class might handle this, or you can ensure it)

\usepackage{booktabs}       % professional-quality tables
\usepackage{amsfonts}       % blackboard math symbols
\usepackage{nicefrac}       % compact symbols for 1/2, etc.
\usepackage{microtype}      % microtypography
\usepackage{newunicodechar}
\newunicodechar{～}{\textasciitilde{}}
\usepackage{xcolor}         % colors
     
\usepackage{make cell}           % 
\usepackage{colortbl}

\usepackage{amssymb}
\usepackage{mathtools}      % mathtools loads amsmath
\usepackage{amsthm}         % If you need theorem environments

\usepackage{enumitem}
\usepackage{algorithm}
\usepackage{algpseudocode}
\usepackage{tabularx}
\usepackage{multirow}
\usepackage{pifont}
\usepackage{fontawesome5}
\usepackage{graphicx}

\usepackage{adjustbox}      % Optional, but in reference. Provides its own \resizebox.
\usepackage{tikz}
\usetikzlibrary{
    arrows.meta,
    shapes.geometric, % For rounded corners and other shapes
    positioning
}
\usepackage{capt-of}
\usepackage[edges]{forest}  % CRUCIAL FOR THE TREE

\usepackage{tcolorbox}
\tcbuselibrary{skins, breakable, most} % Added 

\definecolor{hidden-red}{RGB}{205, 44, 36}
\definecolor{hidden-blue}{RGB}{194,232,247}
\definecolor{hidden-orange}{RGB}{243,202,120}
\definecolor{hidden-green}{RGB}{34,139,34}
\definecolor{hidden-pink}{RGB}{255,245,247}
\definecolor{hidden-black}{RGB}{20,68,106}
\definecolor{purple}{RGB}{144,153,196}
\definecolor{yellow}{RGB}{255,228,123}
\definecolor{hidden-yellow}{RGB}{255,248,203}
\definecolor{tkcolor}{RGB}{224,223,255}
\definecolor{hidden-draw}{RGB}{128,128,128}
\definecolor{darkblue}{rgb}{0, 0.40, 0.75} % 
\definecolor{colorEssential}{HTML}{0D47A1} % A strong, deep blue
\definecolor{colorCommon}{HTML}{1E88E5}    % A standard, clear blue
\definecolor{colorNiche}{HTML}{BDBDBD}      % A neutral gray

\hypersetup{colorlinks=true, citecolor=darkblue, linkcolor=darkblue, urlcolor=darkblue}

\tcbset{
  aibox/.style={
    width=\linewidth,
    top=8pt,
  }
}
\newtcolorbox{AIbox}[2][]{aibox,title=#2,#1}
\tikzset{
  my-box/.style={
    rectangle,
    draw=hidden-black,
    rounded corners,
    text opacity=1,
    minimum height=1.5em,
    minimum width=5em,
    inner sep=2pt,
    align=center,
    fill opacity=.5,
  },
  leaf/.style={
    my-box,
    minimum height=1.5em,
    fill=yellow!32,
    text=black,
    align=left,
    font=\normalsize,
    inner xsep=5pt,
    inner ysep=4pt,
    text width=45em, % This will be scaled by resizebox
  },
  leaf2/.style={
    my-box,
    minimum height=1.5em,
    fill=purple!27,
    text=black,
    align=left,
    font=\normalsize,
    inner xsep=5pt,
    inner ysep=4pt,
  },
  leaf3/.style={
    my-box,
    minimum height=1.5em,
    fill=hidden-blue!57,
    text=black,
    align=left,
    font=\normalsize,
    inner xsep=5pt,
    inner ysep=4pt,
  },
  leaf4/.style={
    my-box,
    minimum height=1.5em,
    fill=green!14,
    text=black,
    align=left,
    font=\normalsize,
    inner xsep=5pt,
    inner ysep=4pt,
    },
    leaf5/.style={
        my-box,
        minimum height=1.5em,
        fill=orange!16,
        text=black,
        align=left,
        font=\normalsize,
        inner xsep=5pt,
        inner ysep=4pt,
    },
}

\tcbset{
  takeawaybox/.style={
    width=\linewidth,
    top=8pt,
    bottom=4pt,
    colback=hidden-yellow, % Different background color
    colframe=black,
    colbacktitle=black,    % Title background color
    fonttitle=\bfseries,
    enhanced,
    center title,
    attach boxed title to top left={yshift=-0.1in,xshift=0.15in},
    boxed title style={boxrule=0pt,colframe=white,},
  }
}
\newtcolorbox{TakeawayBox}[2][]{takeawaybox,title=#2,#1}

\usepackage{tcolorbox}
\newcommand{\questionbox}[1]{%
    \begin{tcolorbox}[colframe=black!60, colback=blue!5, boxrule=1pt, arc=4mm]
        \textbf{\textit{#1}}
    \end{tcolorbox}
}

\NewDocumentCommand{\heng}
{ mO{} }{\textcolor{red}{\textsuperscript{\textit{Heng}}\textsf{\textbf{\small[#1]}}}}

\title{Thinking with Images for Multimodal Reasoning: Foundations, Methods, and Future Frontiers}
\author{%
    Zhaochen Su$^{1}$, \ Peng Xia$^{2}$, \ Hangyu Guo$^{1}$, \ Zhenhua Liu$^{1}$,  \ Yan Ma, \ Xiaoye Qu, \  \\ 
       \textbf{Jiaqi Liu$^{2}$,} \ \textbf{Yanshu Li$^{1}$,} \ \textbf{Kaide Zeng$^{2}$,} \ \textbf{Zhengyuan Yang$^{3}$,} \ \textbf{Linjie Li$^{3}$,} \\ \textbf{Yu Cheng$^{4}$,} \ \textbf{Heng Ji$^{5}$,} \  \textbf{Junxian He$^{1}$,} \ \textbf{Yi R. (May) Fung$^{1}$} \\
    $^{1}$The Hong Kong University of Science and Technology \\ $^{2}$UNC-Chapel Hill, $^{3}$Microsoft, 
    $^{4}$The Chinese University of Hong Kong, $^{5}$UIUC 
    \vspace{0.4em}
}

\begin{document}

\maketitle

% \begin{center}
% \vspace{-26pt}
%   \textbf{Code Page:} \href{https://github.com/zhaochen0110/Awesome_Think_With_Images}{\textcolor{blue}{https://github.com/CSfufu/Revisual-R1}}
% \end{center}

% \begin{center}
% \begin{tabular}{rll}
%     \faGithub & \textbf{\small{Page}} & \url{https://github.com/zhaochen0110/Awesome_Think_With_Images}\\
% \end{tabular}
% \end{center}

\vspace{-20pt}

\begin{abstract}
Recent progress in multimodal reasoning has been significantly advanced by textual Chain-of-Thought (CoT), a paradigm where models conduct reasoning within language. This text-centric approach, however, treats vision as a static, initial context, creating a fundamental semantic gap between rich perceptual data and discrete symbolic thought. Human cognition often transcends language, utilizing vision as a dynamic mental sketchpad. A similar evolution is now unfolding in AI, marking a fundamental paradigm shift from models that merely think about images to those that can truly think with images.
This emerging paradigm is characterized by models leveraging visual information as intermediate steps in their thought process, transforming vision from a passive input into a dynamic, manipulable cognitive workspace.
In this survey, we chart this evolution of intelligence along a spectrum of increasing cognitive autonomy, which unfolds across \textbf{three} key stages: from external tool exploration, through programmatic manipulation, to intrinsic imagination. To structure this rapidly evolving field, our survey makes \textbf{four} key contributions. (1) We establish the foundational principles of the ``Thinking with Images'' paradigm and its three-stage framework. (2) We provide a comprehensive review of the core methods that characterize each stage of this roadmap. (3) We analyze the critical landscape of evaluation benchmarks and transformative applications. (4) We identify significant challenges and outline promising future directions. Through this structured overview, we aim to offer a clear roadmap for future research towards more powerful and human-aligned multimodal AI.\footnote{We maintain a real-time GitHub repository tracking progress at: \url{https://github.com/zhaochen0110/Awesome_Think_With_Images}.}
\end{abstract}

\vspace{-20pt}
\begin{center}   

\centering
\includegraphics[width=1.0\linewidth]{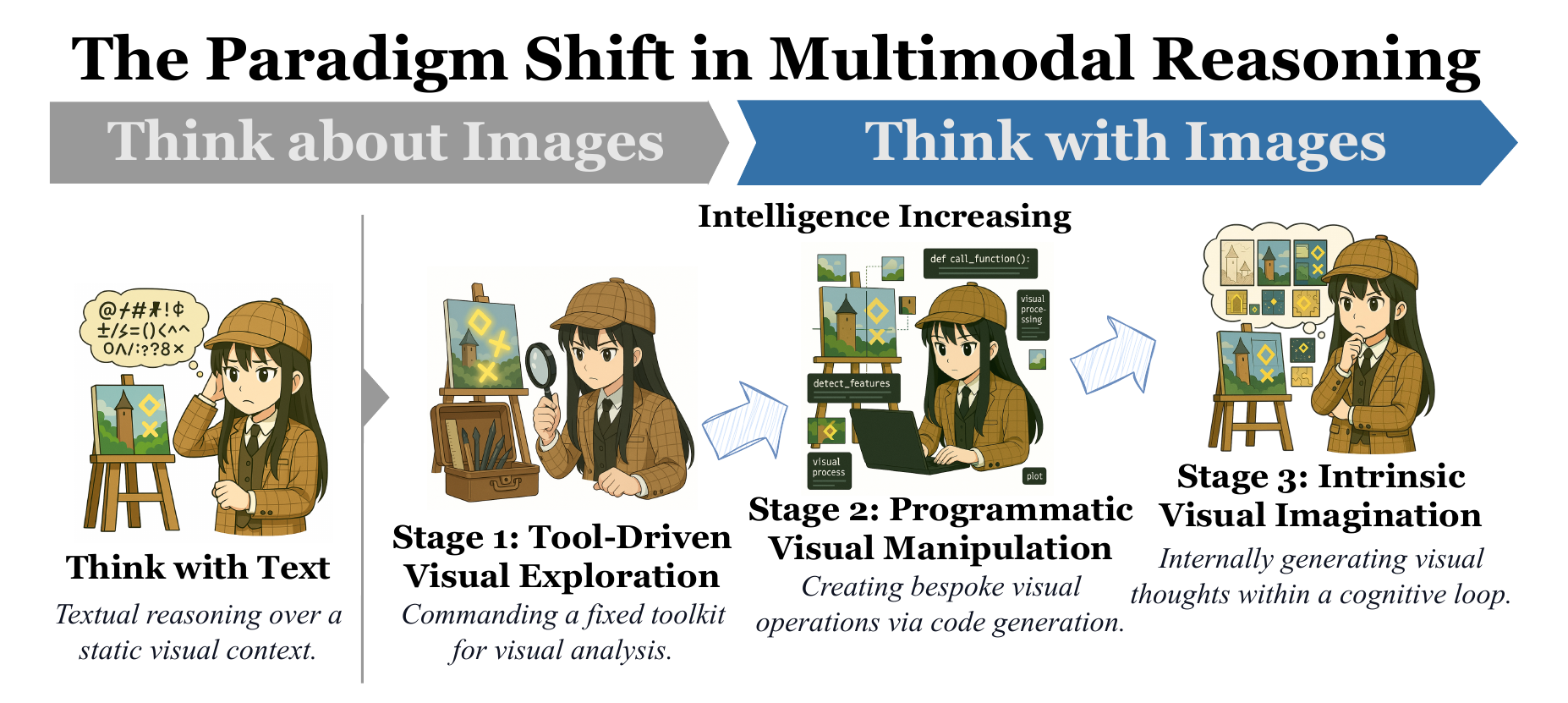}
% \vspace{-26pt}
% \captionof{figure}{The paradigm shift from Thinking about Images'' to ``Thinking with Images'', an evolution that transforms vision from a static input into a dynamic and manipulable cognitive workspace. Each stage marks a significant leap in cognitive autonomy, moving AI towards a more integrated and powerful form of multimodal intelligence.}
\label{main}
\end{center}

\vspace{-10pt}

\tableofcontents
\clearpage

\clearpage
\section{Introduction}
\label{sec:intro}

Large Multimodal Models (LMMs) have recently marked a pivotal moment in artificial intelligence, demonstrating remarkable success in comprehending and generating multimodal content~\citep{team2023gemini, liu2023visual, wang2024qwen2, chen2025advancing}. This progress has fundamentally reshaped AI's ability to bridge the cognitive gap between visual perception and linguistic abstraction. A key catalyst for this first wave of advancement has been the adaptation of language-centric reasoning mechanisms, most notably the Chain-of-Thought (CoT) paradigm~\citep{wei2022chain, kojima2022large}. By decomposing complex problems into a sequence of textual reasoning steps, CoT has significantly elevated LMM performance on a wide array of multimodal tasks, including visual question answering~\citep{zhang2023multimodal,he2025mmboundaryadvancingmllmknowledge,shen2025satori}, visually-grounded mathematical problem-solving~\citep{lu2023mathvista}, and intricate narrative generation~\citep{wu2024next}.

In this established paradigm, which we term ``Thinking about Images'', the visual modality primarily serves as a static, initial context. LMM first ``sees'' an image, encodes it into a fixed set of features, and then operates exclusively within the textual domain to conduct its reasoning. The image is the premise, but language is the exclusive medium of thought. While powerful, this text-centric approach has exposed a fundamental limitation: the semantic gap between the rich, continuous, and often ambiguous nature of the visual world and the discrete, symbolic structure of language~\citep{li2022blip}. This initial, one-time encoding flattens the visual world into a static representation, creating a critical information bottleneck. Consequently, models often falter on tasks that demand deeper, iterative visual engagement, such as complex physical reasoning~\citep{Balazadeh2024Synthetic}, precise spatial manipulation~\citep{gupta2023visual}, or long-horizon planning in interactive environments~\citep{pahuja2025explorer,ragen}.

Now, a new revolution is quietly unfolding in multimodal reasoning. Models are evolving beyond merely thinking about images with text, towards a new paradigm where they can truly think with images. This represents a fundamental shift in the cognitive role of vision: from a passive, fixed input to a dynamic, manipulable cognitive workspace. Much like a human using a sketchpad, models are now being empowered to actively query, modify, and even generate new visual information as integral, intermediate steps within their reasoning process. This ability to form a ``visual chain of thought'' is not a mere extension of textual CoT, but a revolutionary step towards a more holistic and human-like form of cognition~\citep{larkin1987diagrams}. We argue that this emerging paradigm, ``Thinking with Images'', constitutes the next frontier for multimodal AI. It is defined by its core principle: \textit{\textbf{to utilize visual representations as a form of manipulable and verifiable thought, thereby empowering models to actively see, manipulate, and reason with visual information as intermediate steps in a cognitive process.}} This survey provides the first comprehensive and systematic overview of this nascent and rapidly accelerating field.

To structure this rapidly evolving landscape, we propose a conceptual framework that charts the progression of this paradigm through three stages of increasing cognitive autonomy. These stages represent the different mechanisms (the ``How'') through which a model can achieve its visual reasoning goals (the ``Why''): from acting as a ``commander'' that orchestrates external visual tools, to evolving into a ``visual programmer'' that generates code for tailored operations, and ultimately becoming a ``visual thinker'' capable of intrinsic imagination and simulation. This three-stage evolution will be detailed in Section~\ref{sec:foundations}. The proliferation of methods across these stages naturally raises a critical question that this survey seeks to answer:

\questionbox{How are Large Multimodal Models evolving to ``Thinking with Images'' through these advancing stages of cognitive autonomy, and what are the foundational methods, evaluations, applications, and challenges defining this new paradigm?}

This survey is organized to systematically address this question, following the structure of our proposed taxonomy illustrated in Figure~\ref{fig:think-with-image-survey}. We begin by establishing the foundations of this paradigm in \S\ref{sec:foundations}. We then delve into the methodologies of each three stages in \S\ref{sec:stage1}, \S\ref{sec:stage2}, and \S\ref{sec:stage3}, respectively. Following this, we review the critical landscape of evaluation benchmarks and implementation frameworks in \S\ref{sec:evaluation}. We then explore the transformative applications of this paradigm in \S\ref{sec:application}, and discuss prominent challenges and future directions in \S\ref{sec:challenges}. By providing a clear taxonomy and a forward-looking perspective, we aim to not only review existing knowledge but also to inspire future research towards building more powerful, intuitive, and truly multimodal AI.

\tikzstyle{my-box}=[
    rectangle,
    draw=hidden-draw,
    rounded corners,
    text opacity=1,
    minimum height=1.5em,
    minimum width=5em,
    inner sep=2pt,
    align=center,
    fill opacity=.5,
]

\tikzstyle{level1-style}=[
    my-box,
    fill=gray!5,
    text=black,
    font=\scriptsize,
    inner xsep=2pt,
    inner ysep=3pt,
]

\tikzstyle{level2-style}=[
    my-box,
    fill=gray!10,
    text=black,
    font=\scriptsize,
    inner xsep=2pt,
    inner ysep=3pt,
]

\tikzstyle{level3-style}=[
    my-box,
    fill=gray!15,
    text=black,
    font=\scriptsize,
    inner xsep=2pt,
    inner ysep=3pt,
]

\tikzstyle{leaf}=[
    my-box, 
    minimum height=1.5em,
    fill=yellow!32, 
    text=black, 
    align=left,
    font=\scriptsize,
    inner xsep=2pt,
    inner ysep=4pt,
]
\tikzstyle{leaf2}=[
    my-box, 
    minimum height=1.5em,
    fill=purple!27, 
    text=black, 
    align=left,
    font=\scriptsize,
    inner xsep=2pt,
    inner ysep=4pt,
]
\tikzstyle{leaf3}=[
    my-box, 
    minimum height=1.5em,
    fill=hidden-blue!57, 
    text=black, 
    align=left,
    font=\scriptsize,
    inner xsep=2pt,
    inner ysep=4pt,
]
\tikzstyle{leaf4}=[
    my-box, 
    minimum height=1.5em,
    fill=green!14, 
    text=black, 
    align=left,
    font=\scriptsize,
    inner xsep=2pt,
    inner ysep=4pt,
]
\tikzstyle{leaf5}=[
    my-box, 
    minimum height=1.5em,
    fill=orange!16, 
    text=black, 
    align=left,
    font=\scriptsize,
    inner xsep=2pt,
    inner ysep=4pt,
]

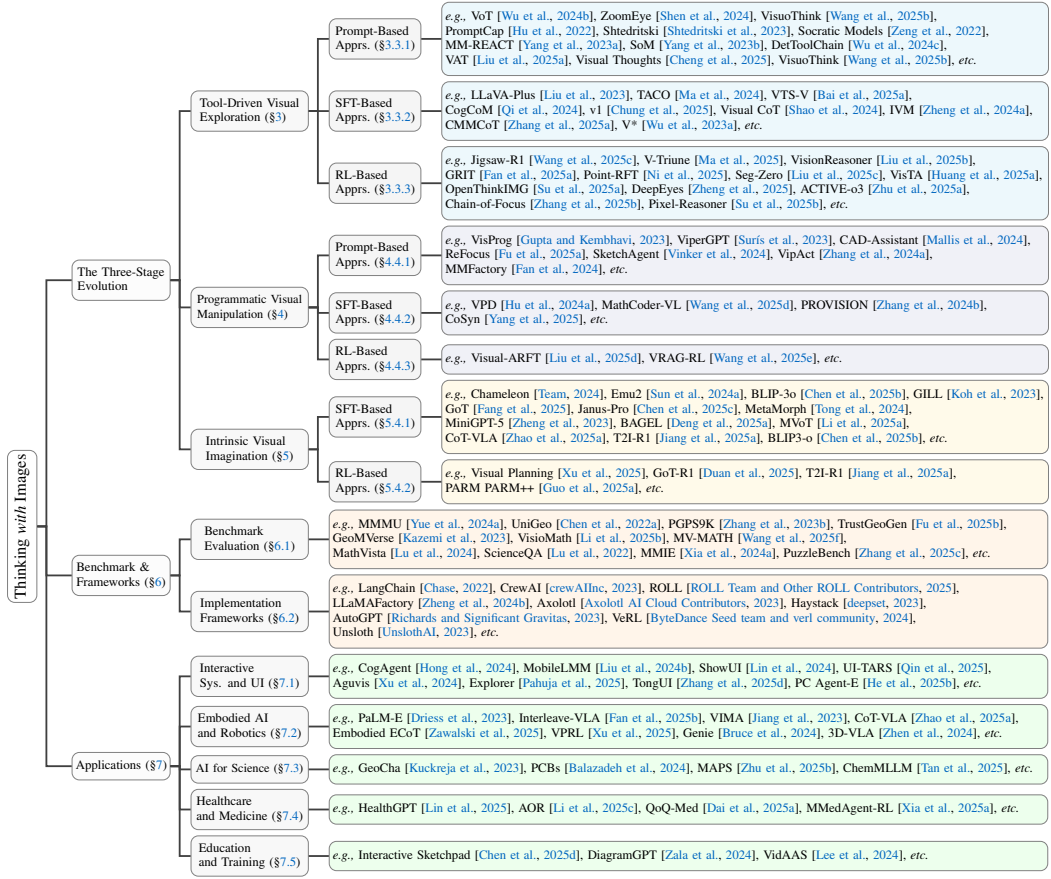
\begin{figure*}[t]
    \centering
    \resizebox{\textwidth}{!}{
        \begin{forest}
            forked edges,
            for tree={
                grow=east,
                reversed=true,
                anchor=base west,
                parent anchor=east,
                child anchor=west,
                base=left,
                font=\small,
                rectangle,
                draw=hidden-draw,
                rounded corners,
                align=left,
                minimum width=4em,
                edge+={darkgray, line width=1pt},
                s sep=3pt,
                inner xsep=2pt,
                inner ysep=3pt,
                ver/.style={rotate=90, child anchor=north, parent anchor=south, anchor=center},
            },
            where level=1{text width=4.95em,font=\scriptsize,level1-style}{},
            where level=2{text width=6.0em,font=\scriptsize,level2-style}{},
            where level=3{text width=4.3em,font=\scriptsize,level3-style}{},
            where level=4{text width=6.1em,font=\scriptsize,}{},
            [
                Thinking \textit{with} Images, ver
                [
                     The Three-Stage \\ Evolution
                    [
                        Tool-Driven Visual \\ Exploration~(\S\ref{sec:stage1})
                        [
                            Prompt-Based \\ Apprs.~(\S\ref{sec:stage1-Approaches-Prompt-Based})
                            [
                                \textit{e.g.,}
                                VoT~\citep{wu2024mind}{,} 
                                ZoomEye~\citep{shen2024zoomeye}{,} 
                                VisuoThink~\citep{wang2025visuothink}{,} \\
                                PromptCap~\citep{hu2022promptcap}{,}
                                Shtedritski~\citep{shtedritski2023does}{,} 
                                Socratic Models~\citep{zeng2022socratic}{,} \\
                                MM-REACT~\citep{yang2023mmreact}{,} 
                                SoM~\citep{yang2023setofmark}{,}
                                DetToolChain~\citep{wu2024dettoolchain}{,} \\
                                VAT~\citep{liu2025visual}{,} 
                                Visual Thoughts~\citep{cheng2025visual}{,}
                                VisuoThink~\citep{wang2025visuothink}{,} 
                                \textit{etc.}
                                , leaf3, text width=32.7em
                            ]
                        ]
                        [
                            SFT-Based \\ Apprs.~(\S\ref{sec:stage1-Approaches-SFT-Based})
                            [
                                \textit{e.g.,}
                                LLaVA-Plus~\citep{liu2023llavaplus}{,}
                                TACO~\citep{ma2024tacolearningmultimodalaction}{,}
                                VTS-V~\citep{bai2025multi}{,} \\
                                CogCoM~\citep{qi2024cogcom}{,} 
                                v1~\citep{chung2025don}{,} 
                                Visual CoT~\citep{shao2024visual}{,}
                                IVM~\citep{zheng2024instruction}{,}\\
                                CMMCoT~\citep{zhang2025cmmcot}{,} 
                                V*~\citep{wu2023multimodal}{,}
                                \textit{etc.}
                                , leaf3, text width=32.7em
                            ]
                        ]
                        [
                            RL-Based \\ Apprs.~(\S\ref{sec:stage1-Approaches-RL-Based})
                            [
                                \textit{e.g.,}
                                Jigsaw-R1~\citep{wang2025jigsaw}{,}
                                V-Triune~\citep{ma2025one}{,}
                                VisionReasoner~\citep{liu2025visionreasoner}{,} \\
                                GRIT~\citep{fan2025gritteachingmllmsthink}{,}
                                Point-RFT~\citep{ni2025pointrftimprovingmultimodalreasoning}{,}
                                Seg-Zero~\citep{liu2025segzero}{,}
                                VisTA~\citep{huang2025visualtoolagent}{,} \\
                                OpenThinkIMG~\citep{su2025openthinkimg}{,}
                                DeepEyes~\citep{zheng2025deepeyesincentivizingthinkingimages}{,}
                                ACTIVE-o3~\citep{zhu2025activeo3}{,} \\
                                Chain-of-Focus~\citep{zhang2025chainoffocusadaptivevisualsearch}{,}
                                Pixel-Reasoner~\citep{pixelreasoner}{,}
                                \textit{etc.}
                                , leaf3, text width=32.7em
                            ]
                        ]
                    ]
                    [
                        Programmatic Visual \\ Manipulation~(\S\ref{sec:stage2}), font=\fontsize{6.7}{8}\selectfont,
                        [
                            Prompt-Based \\ Apprs.~(\S\ref{sec:stage2-Approaches-Prompt-Based})
                            [
                                \textit{e.g.,}
                                VisProg~\citep{gupta2023visual}{,}
                                ViperGPT~\citep{surismenon2023vipergpt}{,}
                                CAD-Assistant~\citep{mallis2024cad}{,}\\
                                ReFocus~\citep{fu2025refocus}{,}
                                SketchAgent~\citep{vinker2024sketchagent}{,}
                                VipAct~\citep{zhang2024vipact}{,} \\
                                MMFactory~\citep{fan2024mmfactory}{,}
                                \textit{etc.}
                                , leaf2, text width=32.7em
                            ]
                        ]
                        [
                            SFT-Based \\ Apprs.~(\S\ref{sec:stage2-Approaches-SFT-Based})
                            [
                                \textit{e.g.,}
                                VPD~\citep{hu2024programdistillation}{,}
                                MathCoder-VL~\citep{wang2025mathcoder}{,}
                                PROVISION~\citep{zhang2024provision}{,} \\
                                CoSyn~\citep{yang2025scaling}{,}
                                \textit{etc.}
                                , leaf2, text width=32.7em
                            ]
                        ]
                        [
                            RL-Based \\ Apprs.~(\S\ref{sec:stage2-Approaches-RL-Based})
                            [
                                \textit{e.g.,}
                                Visual-ARFT~\citep{liu2025VisualARFT}{,}
                                VRAG-RL~\citep{wang2025vrag}{,}
                                \textit{etc.}
                                , leaf2, text width=32.7em
                            ]
                        ]
                    ]
                    [
                        Intrinsic Visual \\ Imagination~(\S\ref{sec:stage3})
                        [
                            SFT-Based \\ Apprs.~(\S\ref{sec:stage3-Approaches-SFT-Based})
                            [
                                \textit{e.g.,}
                                Chameleon~\citep{team2024chameleon}{,} 
                                Emu2~\citep{sun2024generative}{,}
                                BLIP-3o~\citep{chen2025blip3}{,}
                                GILL~\citep{koh2023generating}{,} \\
                                GoT~\citep{fang2025got}{,}
                                Janus-Pro~\citep{chen2025janus}{,}
                                MetaMorph~\citep{tong2024metamorph}{,}\\
                                MiniGPT-5~\citep{zheng2023minigpt}{,} 
                                BAGEL~\citep{deng2025bagel}{,}
                                MVoT~\citep{li2025imaginereasoningspacemultimodal}{,} \\
                                CoT-VLA~\citep{zhao2025cot}{,}
                                T2I-R1~\citep{jiang2025t2i}{,} 
                                BLIP3-o~\citep{chen2025blip3}{,}
                                \textit{etc.}
                                , leaf, text width=32.7em
                            ]
                        ]
                        [
                            RL-Based \\ Apprs.~(\S\ref{sec:stage3-Approaches-RL-Based})
                            [
                                \textit{e.g.,}
                                Visual Planning~\citep{xu2025visualplanningletsthink}{,}
                                GoT-R1~\citep{duan2025got}{,}
                                T2I-R1~\citep{jiang2025t2i}{,} \\
                                PARM PARM++~\citep{guo2025can}{,}
                                \textit{etc.}
                                , leaf, text width=32.7em
                            ]
                        ]
                    ]
                ]
                [
                    Benchmark \& \\ Frameworks~(\S\ref{sec:evaluation})
                    [
                        Benchmark \\ Evaluation~(\S\ref{ssec:bench})
                        [
                            \textit{e.g.,}
                            MMMU~\citep{yue-mmmu-cvpr-2024}{,}
                            UniGeo~\citep{chen-unigeo-emnlp-2022}{,}
                            PGPS9K~\citep{liang-PGPS9K-ijcai-2023}{,}
                            TrustGeoGen~\citep{fu-Trustgeogen-arxib-2025}{,} \\
                            GeoMVerse~\citep{kazemi-geomverse-arxiv-2023}{,}
                            VisioMath~\citep{li-visiomath-arxiv-2025}{,}
                            MV-MATH~\citep{wang-mvmath-arxiv-2025}{,} \\
                            MathVista~\citep{lu-mathvista-iclr-2024}{,}
                            ScienceQA~\citep{lu-scienceqa-nips-2022}{,}
                            MMIE~\citep{xia2024mmie}{,}
                            PuzzleBench~\citep{zhang-PuzzleBench-arxiv-2025}{,}
                            \textit{etc.}
                            , leaf5, text width=38.8em
                        ]
                    ]
                    [
                        Implementation \\Frameworks~(\S\ref{ssec:framework})
                        [
                            \textit{e.g.,}
                            LangChain~\citep{langchain}{,} 
                            CrewAI~\citep{crewai}{,}
                            ROLL~\citep{roll2025alibaba}{,} \\
                            LLaMAFactory~\citep{LlamFactory}{,}
                            Axolotl~\citep{Axolotl}{,}
                            Haystack~\citep{Haystack}{,}\\
                            AutoGPT~\citep{AutoGPT}{,}
                            VeRL~\citep{verl_software_2024}{,}\\
                            Unsloth~\citep{Unsloth}{,}
                            \textit{etc.}
                            , leaf5, text width=38.8em
                        ]
                    ]
                ]
                [
                    Applications~(\S\ref{sec:application})
                    [
                        Interactive \\ Sys. and UI~(\S\ref{ssec:app_interactive_systems}), align=left
                        [
                            \textit{e.g.,}
                            CogAgent~\citep{Hong2024CogAgent}{,}
                            MobileLMM~\citep{liu2024mobilellm}{,}
                            ShowUI~\citep{lin2024showui}{,}
                            UI-TARS~\citep{qin2025ui}{,}\\
                            Aguvis~\citep{xu2024aguvis}{,}
                            Explorer~\citep{pahuja2025explorer}{,}
                            TongUI~\citep{zhang2025tongui}{,}
                            PC Agent-E~\citep{he2025efficient}{,}
                            \textit{etc.}
                            , leaf4, text width=38.8em
                        ]
                    ]
                    [
                        Embodied AI\\and Robotics~(\S\ref{ssec:app_embodied_ai}), align=left
                        [
                            \textit{e.g.,}
                            PaLM-E~\citep{Driess2023PaLME}{,}
                            Interleave-VLA~\citep{Fan2025InterleaveVLAa}{,}
                            VIMA~\citep{Jiang2023VIMA}{,}
                            CoT-VLA~\citep{zhao2025cot}{,} \\
                            Embodied ECoT~\citep{Zawalski2025Robotic}{,}
                            VPRL~\citep{xu2025visualplanningletsthink}{,}
                            Genie~\citep{Bruce2024Genie}{,}
                            3D-VLA~\citep{Zhen20243DVLA}{,}
                            \textit{etc.}
                            , leaf4, text width=38.8em
                        ]
                    ]
                    [
                        AI for Science~(\S\ref{ssec:app_ai_for_science}), align=left, font=\fontsize{6.7}{8}\selectfont,
                        % AI for Sci.~(\S\ref{ssec:app_ai_for_science}), align=left,
                        [
                            \textit{e.g.,}
                            GeoCha~\citep{Kuckreja2023GeoChat}{,}
                            PCBs~\citep{Balazadeh2024Synthetic}{,}
                            MAPS~\citep{zhu2025maps}{,}
                            ChemMLLM~\citep{tan2025chemmllm}{,}
                            \textit{etc.}
                            , leaf4, text width=38.8em
                        ]
                    ]
                    [
                        Healthcare \\and Medicine~(\S\ref{ssec:app_healthcare}), align=left
                        [
                            \textit{e.g.,}
                            HealthGPT~\citep{lin2025healthgpt}{,}
                            AOR~\citep{li2025aor}{,}
                            QoQ-Med~\citep{dai2025qoq}{,}
                            MMedAgent-RL~\citep{xia2025mmedagent}{,}
                            \textit{etc.}
                            , leaf4, text width=38.8em
                        ]
                    ]
                    [
                        Education \\and Training~(\S\ref{ssec:app_education}), align=left
                        [
                            \textit{e.g.,}
                            Interactive Sketchpad~\citep{chen2025interactive}{,} DiagramGPT~\citep{Zala2024DiagrammerGPT}{,}
                            VidAAS~\citep{Lee2024see}{,}
                            \textit{etc.}
                            , leaf4, text width=38.8em
                        ]
                    ]
                ]
            ]
        \end{forest}
    }
    \caption{The taxonomy of the ``Thinking with Images'' paradigm, organized into three main branches: (1) core methodologies, evolving through Tool-Driven Visual Exploration, Programmatic Visual Manipulation, and Intrinsic Visual Imagination; (2) supporting benchmarks and frameworks; and (3) key applications. Representative works are listed on the leaves.
    }
    \vspace{-10pt}
    \label{fig:think-with-image-survey}
\end{figure*}

\subsection{The Position of Our Survey}
\paragraph{How our focus differs from prior surveys.} Earlier reviews have laid essential groundwork for understanding LMMs, yet they share a common perspective: vision is mainly treated as context while language remains the primary vehicle for reasoning.  
General‐scope surveys such as \citet{Yin_2024}, \citet{zhang2024vision}, and \citet{wu2023multimodal} catalogue architectures, pre‐training corpora, and evaluation protocols, but only briefly touch on how a model might use visual information once it has been encoded.  
\citet{xie2024large} extends the discussion to agent settings, yet still emphasises tool calling rather than internal visual cognition.  
Domain‐specific reviews on mathematical reasoning \citep{yan2024survey}, hallucination~\citep{liu2024survey}, and benchmark~\citep{li2025benchmark} delve deeper into task details but inherit the same text‐centric framing. Very recent work has begun to consider explicit reasoning mechanisms.  
\citet{wang2025multimodal} surveys multimodal CoT prompting, and \citet{li2025perception} discusses perception–reason–plan pipelines.  
Even here, the visual modality is largely passive: models describe or annotate an image once, then proceed with textual deliberation.

\vspace{-5pt}
\paragraph{Our position.}
This survey charts the paradigm shift from ``Thinking about Images'' to ``Thinking with Images''. We are the first to systematically categorize the mechanisms that enable this evolution, where the image is transformed from a static premise into a dynamic, manipulable cognitive workspace. We structure our review along a trajectory of increasing cognitive autonomy, detailing how models are evolving from (1) orchestrating tools for exploration, to (2) programmatically creating visual analyses, and ultimately to (3) intrinsically imagining visual information within a closed cognitive loop. We believe this progression will fundamentally redefine what constitutes a reasoning step, what counts as verifiable evidence, and how true visual intelligence should be measured.

\section{Foundations of the Thinking with Images Paradigm}
\label{sec:foundations}

This section establishes the conceptual foundations of the ``Thinking with Images'' paradigm. We first articulate the cognitive imperative for this shift away from purely text-centric reasoning. We then detail its implementation through a three-stage evolutionary framework, provide a formal definition of its mechanics, and conclude with a balanced analysis of its core advantages and unique challenges.

\subsection{The Cognitive Imperative: Why We Need ``Thinking with Images''?}
\label{ssec:why_we_need}

\begin{figure*}[t]
    \centering
    \includegraphics[width=1\textwidth]{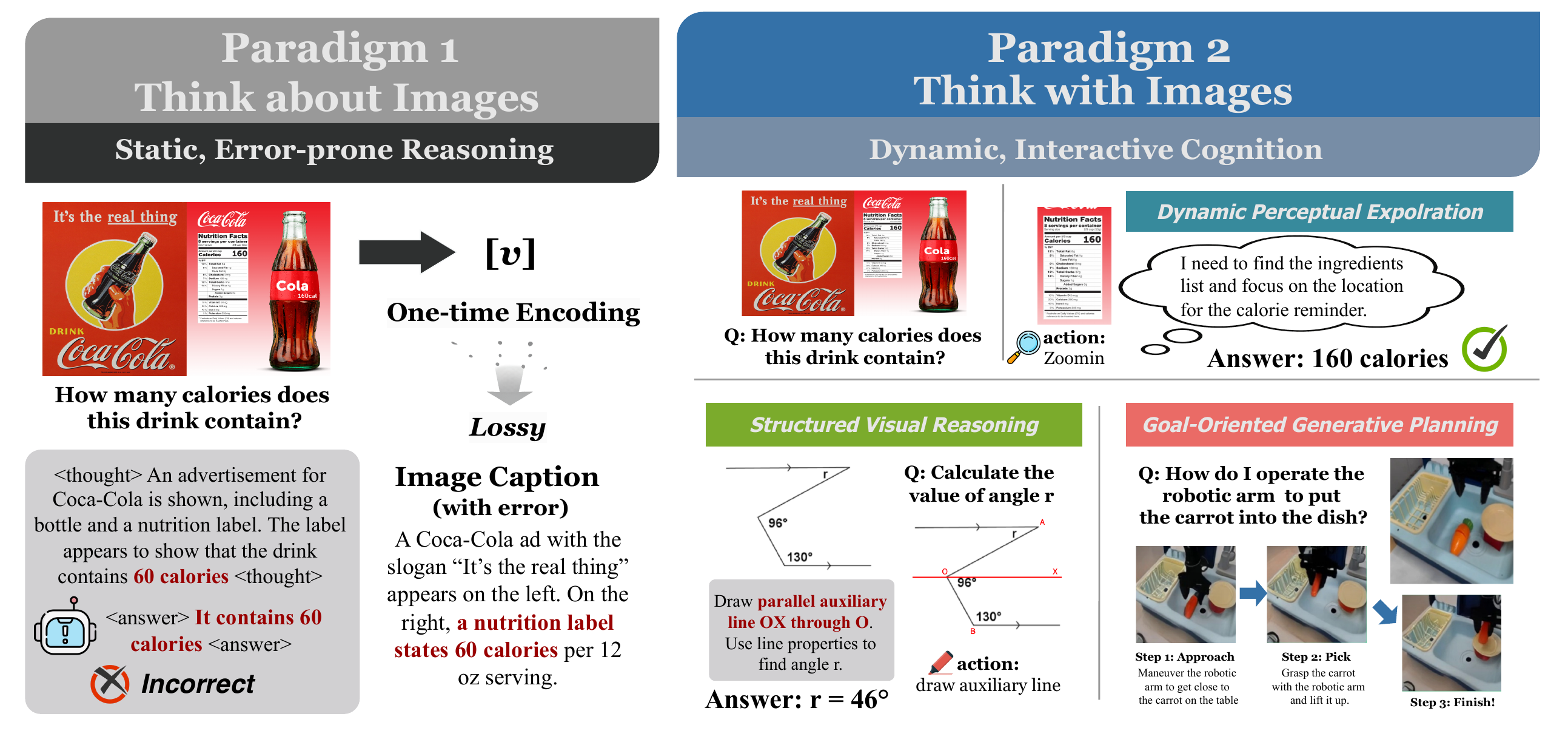}
    \caption{Conceptual comparison of ``Thinking about Images'' versus ``Thinking with Images''. Paradigm 1 shows a failure case of static, error-prone reasoning. Paradigm 2 demonstrates dynamic cognition through three key capabilities: dynamic perceptual exploration, structured visual reasoning, and goal-oriented generative planning. The shift highlights the fundamental transition from static, perceptual input to a dynamic, manipulable cognitive workspace.}
    \label{fig:compare_paradigms}
    % \vspace{-6pt}
\end{figure*}

The prevailing paradigm for multimodal reasoning, where a model first perceives a visual input and then performs subsequent reasoning exclusively through a textual CoT~\citep{wei2022chain, zhang2023multimodal, wang2025multimodal}, has encountered a fundamental ceiling. This approach is constrained by a semantic gap between the high-dimensional, continuous nature of the visual world and the discrete, abstract structure of language. When an image is encoded into a single, static feature vector, the rich, relational structure of the visual world is flattened, making the process of reasoning itself intractable. Consequently, even models with superhuman perception can appear cognitively brittle when facing problems that require more than a static glance. We illustrate this fundamental contrast in Figure~\ref{fig:compare_paradigms} (left), where the ``Thinking about Images'' approach, limited to a single static perception, fails due to lossy, one-time encoding.
The ``Thinking with Images'' paradigm, however, directly overcomes this limitation by transforming the image from a static input into a dynamic and manipulable cognitive workspace. This shift empowers models with a more robust form of cognition, which, as illustrated in Figure~\ref{fig:compare_paradigms} (right), manifests through three key capabilities:

\begin{itemize}[left=2pt,topsep=2pt,itemsep=1pt]
    \item \textbf{Dynamic Perceptual Exploration.} The first capability is to transcend a single, holistic interpretation of an image by engaging in active, iterative inquiry. This is crucial for challenges that require focusing on fine-grained details or understanding object interactions. The model can formalize this dynamic search by generating a new visual reasoning step, for instance by producing a cropped region to ``zoom in'' on an area of interest or by highlighting a particular object. This newly acquired visual evidence enriches the model's immediate understanding, building a more granular and accurate representation of the scene than a single glance could provide. This transforms perception from a one-time encoding into a robust, closed-loop process of observation and targeted examination, allowing the model to resolve visual ambiguities at the pixel level.
    
    \item \textbf{Structured Visual Reasoning.}
    The second capability elevates the image from a mere input into a cognitive scratchpad, fundamentally altering the reasoning process. Consider a complex geometry problem. A purely textual CoT, confined to language, would struggle to solve it, perhaps resorting to lengthy and error-prone algebraic manipulations or failing entirely. In contrast, the ``Thinking with Images'' approach allows the model to generate a new image where a crucial auxiliary line is explicitly rendered. This is a pivotal cognitive leap. When this new image is fed back into the model, previously abstract properties such as newly formed right angles or congruent triangles become perceptually evident features. The reasoning task is thus transformed from difficult symbolic deduction to more robust visual pattern recognition. This process of externalizing thought onto a visual canvas grounds the entire reasoning chain, making it both more powerful and highly interpretable.

    \item \textbf{Goal-Oriented Generative Planning.} Finally, the third capability unlocks goal-oriented generative planning by transforming the model's generative capacity into an engine for simulation. This is essential for tasks requiring physical intuition and forward planning. For example, to determine if a robot arm can grasp an apple without knocking over a nearby glass, a textual CoT must rely on fallible symbolic logic about coordinates and trajectories. In contrast, a model thinking with images can simulate the entire action by generating a hypothetical future frame. In one such imagined future, the model might depict the robot’s gripper colliding with the glass. This generated image is not merely an illustration; it is a form of perceptual self-critique. The visual inconsistency of the collision provides a direct, non-linguistic error signal, forcing the model to re-evaluate its plan. This process of visual simulation is powerful because it outsources validation to the coherence of the physical world itself. A physically implausible plan results in a visually nonsensical image, grounding the model's reasoning in a way impossible by abstract text.
\end{itemize}

% \bigskip
In essence, the three capabilities of active exploration, structured reasoning, and generative planning are not isolated features but interconnected facets of a single, more powerful cognitive process. They collectively transform the visual medium from a static premise to be described into a dynamic workspace for thought. This fundamental shift from passive analysis to active visual engagement enables models to overcome the brittleness of text-centric reasoning, paving the way for a more robust, intuitive, and human-aligned multimodal AI.

\subsection{How Does It Work? The Three-Stage Evolution}
\label{ssec:how_it_works}

The realization of ``Thinking with Images'' signifies an evolutionary journey. We categorize the diverse implementation pathways into three principal stages. These stages reflect a general trend from relying on external systems to developing internalized cognitive functions. To illustrate this evolution, consider the complex task of planning how to rearrange a cluttered room to accommodate a new, larger sofa.

\paragraph{Stage 1: Tool-Driven Visual Exploration.}
In this stage, the model orchestrates a fixed inventory of external visual modules. Its primary role is to act as a planner selecting an appropriate tool for the current sub-task. To address the furniture arrangement problem, the model might first invoke an \texttt{object\_detector} to identify items and a \texttt{distance\_estimator} to measure the available space. As the model's reasoning is grounded in the outputs of these tools, it might conclude that ``the current gap is 1.5 meters, while the new sofa requires 2.0 meters, so it will not fit''. This approach is powerful for targeted data gathering but is limited by the static capabilities of its predefined toolset.

\paragraph{Stage 2: Programmatic Visual Manipulation.}
This stage elevates the model's autonomy by enabling it to function as a visual programmer. Instead of selecting from a finite menu, the model generates executable code to perform a custom visual analysis. Faced with the furniture problem, the model could generate a Python script using a library like \texttt{matplotlib}. This script would create a 2D top-down floor plan of the room, representing each piece of furniture as a shape. The model could programmatically test various new arrangements in this abstract visual space. This method unlocks flexibility and its generated code serves as a transparent, verifiable record of its thought process. The primary constraint remains its reliance on an external execution environment to run the code.

\paragraph{Stage 3: Intrinsic Visual Imagination.}
At the most advanced stage, the model achieves full cognitive autonomy through intrinsic imagination. It transcends the dependency on external execution by leveraging its unified generative architecture to produce a new image as an intermediate reasoning step. To find the optimal room layout, the model could directly generate a new, photorealistic image depicting the room with the furniture rearranged and the new sofa in place. This internally generated image functions as a visual hypothesis or a mental simulation. The model can then feed this image back into itself to critique the new layout, perhaps noticing that a pathway is now blocked. This process enables a seamless and truly integrated form of visual thought within a closed cognitive loop, addressing the architectural bottlenecks of prior stages.

\paragraph{A Note on Non-Linearity.}
These three stages do not represent a strictly linear progression. They are different implementation strategies (the ``How'') that a model can employ to achieve its cognitive goals (the ``Why''). This distinction is critical within our running example of room arrangement. Before performing any complex rearrangement simulation, the model must first answer a simple question: can the new sofa fit through the doorway? A model capable of Stage 3 imagination could attempt to generate a full physical simulation of maneuvering the sofa, but this is computationally expensive and potentially unreliable. A far more efficient and robust strategy is to employ a Stage 1 mechanism by invoking a simple \texttt{measure} tool to get the exact width of the doorway. This illustrates that a model's intelligence lies not only in its peak capability but also in its ability to select the appropriate cognitive tool for the task. Our framework therefore provides a structured way to understand an approach's primary mechanism while acknowledging the field's complex and interconnected nature.

\subsection{What is Thinking with Images? A Formal Definition}
While the three-stage evolution describes the various implementation pathways, a precise, formal definition is necessary to delineate the paradigm's fundamental mechanics. At its core, ``Thinking with Images'' is defined by its treatment of visual information as an operable, intermediate step within the reasoning process itself. This marks a clear departure from prior methods where the image is a fixed, initial condition. We can formalize this distinction by contrasting the two approaches.

\paragraph{Paradigm I: Thinking about Images} In this paradigm, the reasoning process is an autoregressive generation of textual thoughts conditioned on a visual input. Given an LMM with parameters $\Theta_{LMM}$ and an input query $Q$, the image \(I\) is processed once by a visual encoder \(\Phi_V \in \Theta_{LMM} \) to obtain a set of features \(\mathbf{v} = \Phi_V(I)\). These features then serve as a fixed context for a language model. The model generates the reasoning chain by producing text tokens \(x_t\), where the context for each token is the sequence of previously generated text \(x_{<t}\) and the visual features \(\mathbf{v}\).
\begin{equation}
x_t \sim P(\cdot | x_{<t}, \mathbf{v}, Q; \Theta_{LMM})
\label{eq:think_text_revised}
\end{equation}
Here, the visual information \(\mathbf{v}\) acts as an initial condition and is not modified during the process.

\paragraph{Paradigm II: Thinking with Images} This paradigm expands the model's generative capabilities beyond the textual domain. The reasoning process is a sequence of steps that can be either textual or visual. We define the reasoning history at step \(t\) as a state sequence \(S_t = (z_1, \dots, z_{t-1})\). The core action is to generate the next reasoning step \(z_t\) based on this evolving multimodal history. Let $\mathcal{T}_{\text{text}}$ denote the space of all possible textual outputs (e.g., the vocabulary of text tokens) and $\mathcal{I}_{\text{vis}}$ denote the space of all possible intermediate visual artifacts that can be introduced or modified during reasoning. The next step $z_t$ is sampled from the union of these two spaces.

\begin{equation}
    z_t \sim P(\cdot | S_t, I, Q; \Theta_{LMM}) \quad \text{where } z_t \in \mathcal{T}_{\text{text}} \cup \mathcal{I}_{\text{vis}}
    \label{eq:think_images_unified}
\end{equation}
The critical distinction lies in the dynamic nature of state history \(S_t\), which can contain both textual and visual reasoning steps. This formalization supports a broad definition of the paradigm. An intermediate visual step $z_t \in \mathcal{I}_{\text{vis}}$ can manifest in several forms: it can be the output from an external tool, such as the bounding box of a detected object; it might be a visualization generated by code, such as an auxiliary line drawn on a diagram~\citep{gupta2023visual}; or it can be a new image produced intrinsically by the model, such as a predicted future frame~\citep{xu2025visualplanningletsthink}.

\subsection{Unique Challenges}
\label{ssec:unique_challenges}

While ``Thinking with Images'' unlocks profound capabilities, it also introduces a distinct class of challenges that diverge significantly from those in textual CoT. The continuous and dense nature of visual information, which is the source of its power, also creates fundamental obstacles related to efficiency, robustness, and generalization.

\paragraph{Computational Cost: The Explosive Token Economy of Visual Thought.}
``Thinking with Images'' is prohibitively expensive compared to its textual counterpart, limiting the length and complexity of reasoning. While textual reasoning has become more efficient~\citep{sui2025stopoverthinkingsurveyefficient,qu2025survey}, the cost of processing visual information remains orders of magnitude higher. A single image is decomposed into a dense grid of thousands of visual patches, each requiring its own intensive computation. Generating even one intermediate visual step is a major computational event. A multi-step reasoning process that strings these events together results in a compounding multiplicative, often prohibitive, burden. This ``token explosion'' makes exploring long visual reasoning paths practically infeasible with current architectures, creating a hard ceiling on the depth of visual deliberation.

\vspace{-5pt}
\paragraph{Information Density: One Flawed Image versus a Thousand Wrong Words.}
The dense nature of visual information causes errors to propagate in uniquely damaging ways. A flawed assertion in a textual chain can introduce a logical contradiction~\citep{su2024conflictbank,jia2025benchmarking,sun2024surf}. However, a visual error corrupts the underlying perception of reality~\citep{dosovitskiy2020image}. Consider a model using a zoom-in tool to identify a product’s material. If it erroneously focuses on the wooden table underneath, its subsequent analysis, instead of simply failing, may begin a new, internally coherent line of reasoning about a completely irrelevant subject, generating a plausible-sounding analysis of wood grain and varnish sheen. This initial mistake establishes a false ground truth for all subsequent deliberation, creating a foundational error in perception whose interconnected falsehoods poison the entire reasoning process.
%"The dense nature of visual information causes errors to propagate in uniquely damaging ways. While a flawed assertion in a textual chain may introduce a logical contradiction~\citep{su2024conflictbank,jia2025benchmarking,sun2024surf}, a visual error corrupts the underlying perception of reality~\citep{dosovitskiy2020image}. Consider a model using a zoom-in tool to identify a product’s material: if it erroneously focuses on the wooden table underneath, the error transcends irrelevance. Rather than halting or producing incoherent output, the model may generate an internally coherent analysis of the table’s wood grain and varnish sheen—a plausible but entirely unrelated reasoning chain. This mistake establishes a false perceptual foundation, poisoning all subsequent deliberation with interconnected falsehoods."
\vspace{-5pt}

\paragraph{Architectural Divide: The Bottleneck Between Language and Pixels.}
Most current models for ``Thinking with Images'' use a modular design separating vision and language systems~\citep{liu2023visual,bai2025qwen2}. While versatile, this separation creates challenges for iterative visual thinking. The architecture must translate rich spatial information into a sequential format, a process that risks losing fine-grained detail~\citep{team2024chameleon}. True visual thinking ideally requires a dynamic loop, allowing a model to alternate between forming an intention and perceiving its immediate consequences. The current modular design can impede this loop by making the model's perception of its actions indirect. Architectures enabling more direct engagement between the reasoning engine and the visual canvas are therefore a crucial direction for the field.

\vspace{-5pt}

\paragraph{Cross-Task Generalization: A Single Visual Strategy Does Not Fit All.}
The ``Thinking with Images'' paradigm introduces a profound challenge: a single mode of visual thinking is inadequate for the diverse nature of visual tasks. Different problems demand different modes of visual thinking. For example, solving a geometry puzzle might require a constructive strategy of generating new visual elements like lines and circles~\citep{wang2025mathcoder}. Diagnosing a detail-rich image may need an analytical strategy involving a sequence of targeted zoom-ins and comparisons~\citep{zheng2025deepeyesincentivizingthinkingimages}. Navigating a maze could demand a simulative strategy where the model imagines future visual states~\citep{li2025imaginereasoningspacemultimodal}. A model trained with one strategy will fail at tasks requiring another. Developing agents that can maintain a toolkit of diverse visual strategies and learn a meta-policy to select the appropriate one for a given task remains a complex and unsolved challenge.

\begin{figure*}[t]
    \centering
    \includegraphics[width=1\textwidth]{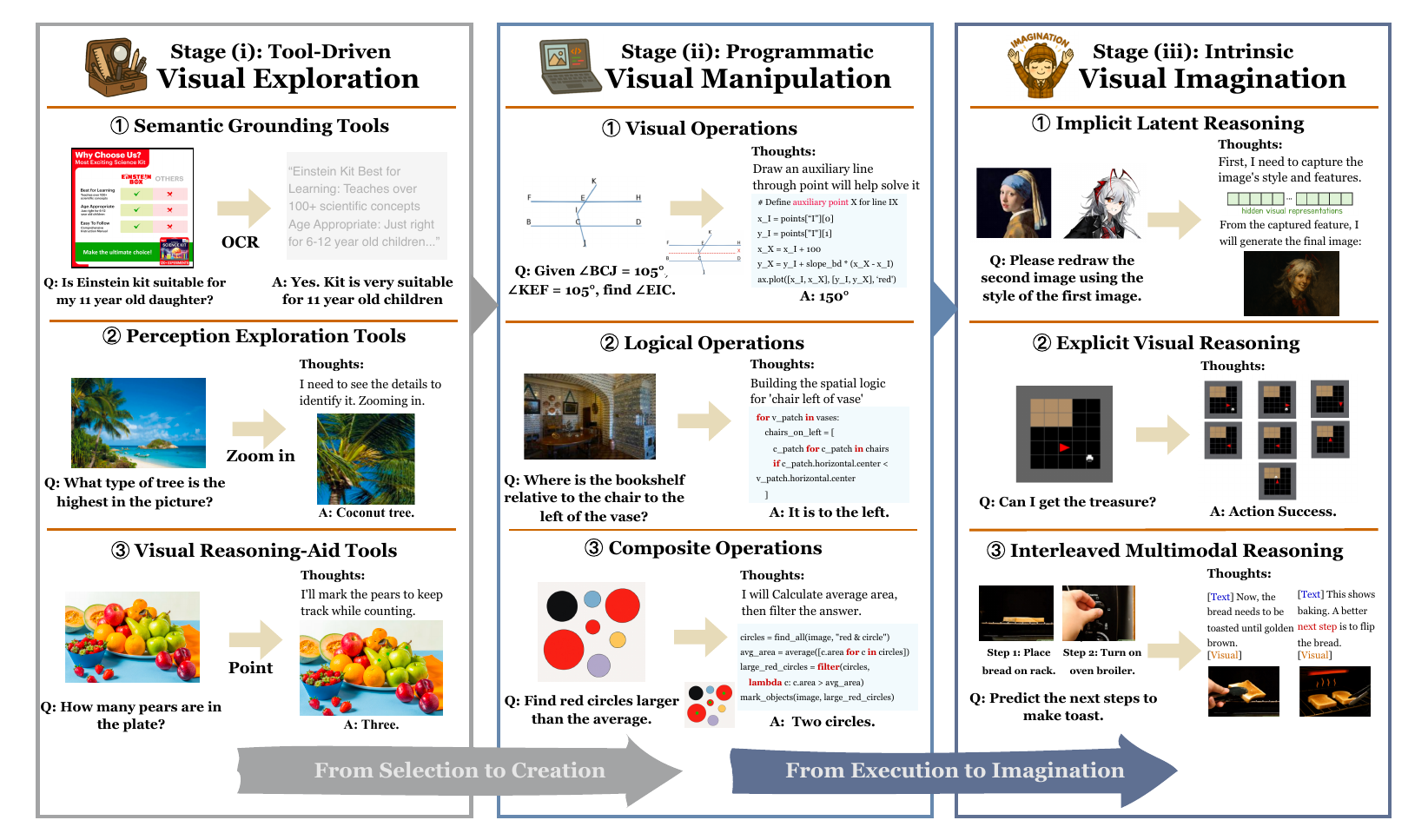}
    \caption{The specific operational categories within each of the three ``Thinking with Images'' paradigms, illustrating the evolution from selection to creation and imagination. \textbf{Stage 1:} The model selects from a fixed toolkit, categorized into semantic grounding, perception exploration, and visual reasoning-aid tools. \textbf{Stage 2:} The model creates customized programs by composing visual, logical, and composite operations. \textbf{Stage 3:} The model generates internal thoughts, which can manifest as implicit latent reasoning, explicit visual reasoning, and interleaved multimodal reasoning.}
    \label{fig:evolve}
    % \vspace{-6pt}
\end{figure*}

\section{Stage 1: Tool-Driven Visual Exploration} 
\label{sec:stage1}

Having established the foundational principles of the ``Thinking with Images'' paradigm, we delve into the mechanics of its three evolutionary stages. This progression, visually charted in Figure~\ref{fig:evolve}, represents a path of increasing cognitive autonomy, moving from the selection of external tools, to the creation of programmatic operations and to intrinsic imagination. Each stage builds upon the last, equipping models with more powerful and integrated forms of visual intelligence.

We begin with the first pivotal step: Tool-Driven Visual Exploration. The model's intelligence lies in acting as a planner or orchestrator. It learns to leverage a predefined suite of external visual tools to conduct its analysis, marking the crucial shift from passive perception to active, tool-driven inquiry. A comprehensive overview of the methods in this stage, their underlying implementation approaches, and their primary application scenarios is provided in Table~\ref{tab:stage1_methods}.

\subsection{Formulation}
\label{sec:stage1-formulation}
In this stage, the LMM acts as a high-level planner that operates on a predefined inventory of external visual tools, denoted as \(\mathcal{T}_{tool} = \{T_1, \dots, T_m\}\). Each tool \(T_i\) performs a specific, encapsulated function. The model's primary role is to determine which tool to use and how to apply it based on its evolving understanding of the problem. The core action at each reasoning step \(t\) is for the LMM's policy to generate a symbolic tool call, \(c_t\). This call specifies the chosen tool from the inventory \(T_{\text{sel}} \in \mathcal{T}\) and the necessary arguments \(a_{\text{tool}}\).
\begin{equation}
c_t \sim P(\cdot | S_t, I, Q; \Theta_{LMM})  \quad \text{where } c_t = (T_{sel}, a_{tool})
\end{equation}
This generated command is then passed to an external tool execution engine \(\mathcal{E}_{\text{tool}}\), which executes the specified operation on the image \(I\). The engine returns an output \(o_t\):
\begin{equation}
o_t = \mathcal{E}_{\text{tool}}(c_t, I) \quad \text{where } o_t \in \mathcal{T}_{\text{text}} \cup \mathcal{I}_{\text{vis}}
\end{equation}
This output \(o_t\), which can be a modified image, a data layer like a segmentation mask, or structured text, is then integrated back into the model's state \(S_{t+1}\) to inform the next reasoning step.

The key characteristic of this stage is that visual operations are selected from a fixed set. This marks a significant departure from the ``Thinking \textit{about} Images'' paradigm, where the visual input is static. Here, the model actively interrogates the visual world through a controlled interface, but its operational vocabulary is finite and predefined. The model's intelligence is demonstrated in its ability to orchestrate this fixed set of tools in a logical sequence.

\subsection{Categories of Visual Tools}
\label{sec:stage1-categories}
The effectiveness of ``Thinking with Images'' hinges on well-curated and diverse tools. In Figure~\ref{fig:evolve}, these can be grouped into three categories, each serving a cognitive purpose in the reasoning chain.

\begin{itemize}[left=2pt,topsep=2pt,itemsep=1pt]
    \item \textbf{Semantic Grounding Tools.} These tools bridge the gap between pixel space and symbolic language by transforming visual content into textual representations. This category includes tools for generating descriptive captions of a scene~\citep{hu2022promptcap} and employing Optical Character Recognition (OCR) to extract embedded text from images, a function central to frameworks like TACO~\citep{ma2024tacolearningmultimodalaction} and CogCoM~\citep{qi2024cogcom}. Their primary role is to provide foundational semantic grounding, converting raw visual data into a discrete, symbolic format that the LMM can directly process and reason about.

    \item \textbf{Perception Exploration Tools.} This category of tools serves to enhance perception, enabling the model to actively examine visual content for finer details that a single, holistic view might miss. The operations occur within the visual domain itself to provide a richer perceptual input. They facilitate this through mechanisms like programmatic cropping or zooming to inspect specific regions with high fidelity~\citep{shen2024zoomeye, zheng2025deepeyesincentivizingthinkingimages}. This category also includes specialized models like object detectors and semantic segmenters, which identify and isolate key elements for a more focused analysis~\citep{wu2024dettoolchain, yang2023setofmark}

    \item \textbf{Visual Reasoning-Aid Tools.} This category of tools directly supports the cognitive process itself, transforming the image into an active workspace for reasoning. The core function is to externalize thought, making abstract logical steps tangible and verifiable on the visual canvas. This is achieved by highlighting critical regions to direct attention, masking irrelevant information to reduce distraction, or drawing new elements like auxiliary lines in a geometry problem or pointers for counting objects~\citep{yang2023setofmark, fu2025refocus}. These actions modify the image not just for better perception, but to actively structure and guide the flow of thought.
\end{itemize}

\newcolumntype{C}[1]{>{\centering\arraybackslash}p{#1}}
\newcolumntype{R}[1]{>{\raggedleft\arraybackslash}p{#1}}

\begin{table*}[!t]
\fontsize{7.5}{9.5}\selectfont
\centering
\setlength{\tabcolsep}{1.5mm} % Adjusted column separation
\renewcommand{\arraystretch}{1.1} 
\resizebox{\linewidth}{!}{
\begin{tabular}{l >{\raggedright\arraybackslash}p{3.0cm} R{1.4cm} C{0.6cm}C{0.6cm}C{0.6cm} C{0.5cm}C{0.5cm}C{0.5cm}}
\toprule
\multirow{2}{*}[-2pt]{\textbf{Methods}} & \multirow{2}{*}[-2pt]{\textbf{Base Model}} & \multirow{2}{*}[-2pt]{\textbf{Data}} & \multicolumn{3}{c}{\textbf{Tool Categories}} & \multicolumn{3}{c}{\textbf{Scenario}} \\
\cmidrule(lr){4-6} \cmidrule(lr){7-9}
& & & \textbf{SG} & \textbf{PE} & \textbf{VR} & \textbf{P} & \textbf{R} & \textbf{G} \\ 
\midrule
\multicolumn{9}{c}{\textit{\textbf{Prompt-Based Approaches}}} \\
\midrule
Socratic M.~\citep{zeng2022socratic} & GPT-3 & -- & \checkmark & & & \checkmark & & \\
PromptCap~\citep{hu2022promptcap} & GPT-3 & -- & \checkmark & & & \checkmark & & \\
MM-REACT~\citep{yang2023mmreact} & GPT-3.5 & -- & \checkmark & \checkmark & & \checkmark & \checkmark & \\
Visual ChatGPT~\citep{wu2023visualchatgpttalkingdrawing} & GPT-3.5 & -- & \checkmark & \checkmark & & \checkmark & \checkmark & \checkmark \\
Shtedritski~\citep{shtedritski2023does} & CLIP & -- & & & \checkmark & \checkmark & & \\
SoM~\citep{yang2023setofmark} & GPT-4V & -- & & \checkmark & \checkmark & \checkmark & & \\
VoT~\citep{wu2024mind} & GPT-4 & -- & & & \checkmark & \checkmark & \checkmark & \\
DetToolChain~\citep{wu2024dettoolchain} & GPT-4V, etc. & -- & & \checkmark & \checkmark & \checkmark & \checkmark & \\
ZoomEye~\citep{shen2024zoomeye} & LLaVA-v1.5-7B, etc. & -- & & \checkmark & & \checkmark & & \\
Chain-of-Spot~\citep{liu2024chain} & LLaVA-v1.5-7B, etc. & -- & & \checkmark & & \checkmark &  & \\
DyFo~\citep{li2025dyfo} & LLaVA-v1.5-7B, etc. & -- & & \checkmark & & \checkmark &  & \\
ViCrop~\citep{zhang2025mllms} & LLaVA-v1.5-7B, etc. & -- & & \checkmark & & \checkmark &  & \\
Visual Thoughts~\citep{cheng2025visual} & LLaVA-v1.5-7B, etc. & -- & \checkmark & \checkmark & & \checkmark & \checkmark & \checkmark \\
VisuoThink~\citep{wang2025visuothink} & GPT-4o, etc. & -- & & \checkmark & \checkmark & \checkmark & \checkmark & \\
VAT~\citep{liu2025visual} & GPT-4o, etc. & -- & & \checkmark & \checkmark & \checkmark & \checkmark & \\
\midrule
\multicolumn{9}{c}{\textit{\textbf{SFT-Based Approaches}}} \\ 
\midrule
LLaVA-Plus~\citep{liu2023llavaplus} & LLaVA-7B & 158K & \checkmark & \checkmark & & \checkmark & \checkmark & \checkmark \\
CogCoM~\citep{qi2024cogcom} & CogVLM-17B & 91K & \checkmark & \checkmark & & \checkmark & \checkmark & \\
Visual CoT~\citep{shao2024visual} & Vicuna-7/13B, etc. & 438K & & \checkmark & & \checkmark & & \\
IVM~\citep{zheng2024instruction} & LLaVA-7B & 334K & & & \checkmark & \checkmark & & \\
VisualReasoner~\citep{cheng2024least} & LLaVA-1.5-7B & 50K & & \checkmark &  & \checkmark & \checkmark & \\
TACO~\citep{ma2024tacolearningmultimodalaction} & LLaMA3-8B, etc. & 293K & \checkmark & \checkmark & & \checkmark & \checkmark & \\
CMMCoT~\citep{zhang2025cmmcot} & Qwen2-VL-7B & 260K & & \checkmark & & \checkmark & \checkmark & \\
v1~\citep{chung2025don} & Qwen2.5-VL-7B & 300K & & \checkmark & \checkmark & \checkmark & \checkmark & \\
V* (SEAL)~\citep{vstar} & LLaVA-7B & 387K & & \checkmark & & \checkmark & \checkmark & \\
VTS-V~\citep{bai2025multi} & Qwen2.5-VL-7B, etc & 301K & &\checkmark & \checkmark & \checkmark & \checkmark &  \\
VGR~\citep{wang2025vgr} & LLaVA-NeXT & 158K & & \checkmark & & \checkmark & \checkmark & \\
UniVG-R1~\citep{bai2025univg} & Qwen2-VL-2B, etc & 94K& &\checkmark & & \checkmark & \checkmark & \\
\midrule
\multicolumn{9}{c}{\textit{\textbf{RL-Based Approaches}}} \\ 
\midrule
Chain-of-Focus~\citep{zhang2025chainoffocusadaptivevisualsearch} & Qwen2.5-VL-7B & 3.7K & & \checkmark & & \checkmark & \checkmark & \\
ACTIVE-o3~\citep{zhu2025activeo3} & Qwen2.5-VL-7B & 1K* & & \checkmark & & \checkmark & \checkmark & \\
DeepEyes~\citep{zheng2025deepeyesincentivizingthinkingimages} & Qwen2.5-VL-7B & 47K & & \checkmark & & \checkmark & \checkmark & \\
GRIT~\citep{fan2025gritteachingmllmsthink} & Qwen2.5-VL-3B, etc. & 0.02K & & \checkmark & & \checkmark & \checkmark & \\
VisTA~\citep{huang2025visualtoolagent} & Qwen2.5-VL-7B & 0.8K* & \checkmark & \checkmark & \checkmark & \checkmark & \checkmark & \\
Jigsaw-R1~\citep{wang2025jigsaw} & Qwen2.5-VL-3/7B, etc. & $\approx$180K & & \checkmark & & \checkmark & \checkmark & \\
Seg-Zero~\citep{liu2025segzero} & Qwen2.5-VL-3B, etc. & 9K & \checkmark & \checkmark & & \checkmark & & \\
PixelThink~\citep{wang2025pixelthink} & Qwen2.5-VL-7B, etc. & 9K & \checkmark & \checkmark & & \checkmark & & \\
VisionReasoner~\citep{liu2025visionreasoner} & Qwen2.5-VL-7B & 66K & \checkmark & \checkmark & & \checkmark & \checkmark & \\
V-Triune~\citep{ma2025one} & Qwen2.5-VL-7/32B & 47K & & \checkmark & & \checkmark & \checkmark & \\
Point-RFT~\citep{ni2025pointrftimprovingmultimodalreasoning} & Qwen2.5-VL-7B & 71K & & \checkmark & & & \checkmark & \\
OpenThinkIMG~\citep{su2025openthinkimg} & Qwen2-VL-2B & 14.5K & & \checkmark & \checkmark & \checkmark & \checkmark & \\
Pixel-Reasoner~\citep{pixelreasoner} & Qwen2.5-VL-7B & 15K & & \checkmark & & \checkmark & \checkmark & \\
VLM-$R^3$~\citep{jiang2025vlm} & Qwen2.5-VL-7B & 12K & & \checkmark & & \checkmark & \checkmark & \\
VILASR~\citep{wu2025reinforcing} & Qwen2.5-VL-7B & 81K & & \checkmark & \checkmark & \checkmark & \checkmark & \\
\bottomrule
\end{tabular}}
\vspace{1mm}

\caption{Methods for Tool-Driven Visual Exploration. This table classifies approaches by their base model and data requirements. It details the categories of tools employed: Semantic Grouding (\textbf{SG}), Perception Exploration (\textbf{PE}), and Visual Reasoning-Aid (\textbf{VR}). It also specifies application scenarios: foundational \textbf{Perception (P)}, complex multi-step \textbf{Reasoning (R)}, and novel \textbf{Generation (G)}. Data amounts marked with * denote per-task requirements.}
\vspace{-15pt}
\label{tab:stage1_methods}
\end{table*}

\subsection{Implementation Approaches}
To equip models with tool orchestration, models must be enabled to act as effective coordinators. This section details the three main methodologies used to instill this capability: prompt-based methods that leverage in-context learning, supervised fine-tuning that teaches procedural competence through examples, and reinforcement learning that enables policy optimization for tool use.

\subsubsection{Prompt-Based Approaches}
\label{sec:stage1-Approaches-Prompt-Based}
Prompt-based methods enable LMMs to coordinate predefined external visual tools without parameter updates. Carefully designed prompts transform static visual inputs into actively explorable workspaces, leveraging the model's in-context learning to facilitate robust problem-solving.

\paragraph{Mediating Collaboration through Language.}
This foundational approach uses language as a universal interface to grant text-only Large Language Models (LLMs) visual reasoning capabilities. Socratic Models~\citep{zeng2022socratic} pioneered this concept by enabling specialized models to collaborate through structured dialogues, where a vision-language model provides perceptual insights for an LLM to process. Building on this, PromptCap~\citep{hu2022promptcap} introduced a more targeted method by developing question-aware captioning models. These models generate task-specific visual descriptions to guide the LLM, rather than generic captions. Similarly, MM-REACT~\citep{yang2023mmreact} demonstrated that a text-only model can act as a central controller, orchestrating a pool of vision experts through prompted instructions to decompose and solve complex visual tasks. These methods establish that multimodal reasoning can emerge from language-mediated coordination, allowing text-only models to function as visual reasoning systems without architectural changes.

\paragraph{Manipulating Input for Perception.}
This category improves the perceptual capabilities of existing LMMs through simple input modifications and systematic exploration. Visual prompt engineering is a technique where drawing a red circle on an image directs a model's attention to specific regions~\citep{shtedritski2023does,zhang2025vlm2benchcloserlookvlms}, with the goal of enhancing zero-shot performance without fine-tuning. Beyond spatial highlighting, Visualization-of-Thought (VoT)~\citep{wu2024mind} enhances spatial reasoning by prompting models to generate textual visualizations like ASCII grids. This process creates an explicit spatial workspace that enables text-only LLMs to solve spatial tasks. Another approach, Visual Abstract Thinking (VAT)~\citep{liu2025visual}, simplifies visual inputs into abstractions like sketches or contours, filtering out distracting noise while preserving essential structural information. For systematic exploration, methods like ZoomEye~\citep{shen2024zoomeye} introduce training-free, tree-based search strategies that allow models to perform human-like zooming operations and explore image hierarchies. Similarly, ViCrop~\citep{zhang2025mllms} and Chain-of-Spot (CoS)~\citep{liu2024chain} leverage the model's internal attention to automatically crop and zoom into relevant regions. VisuoThink~\citep{wang2025visuothink} extends this by integrating multimodal tree search with interleaved reasoning to explore multiple solution paths. These approaches show that simple modifications and structured exploration can enhance model perception without retraining.

\paragraph{Leveraging Specialized Visual Experts.}
This approach integrates specialized visual expert models to achieve high-precision visual understanding. Set-of-Mark (SoM) prompting~\citep{yang2023setofmark} exemplifies this by using a segmentation model to partition an image into regions, which are then overlaid with symbolic marks. This allows a model like GPT-4V to achieve precise visual grounding by referencing these marks in its response. DetToolChain~\citep{wu2024dettoolchain} introduces a prompting paradigm that combines visual processing tools with a structured detection reasoning chain to decompose complex detection tasks. Other work employs Monte Carlo Tree Search~(MCTS) to enable bidirectional interaction between LMMs and visual experts, simulating human-like focus adjustments where expert feedback refines the model's attention~\citep{li2025dyfo}. The Visual Thoughts~\citep{cheng2025visual} framework provides a theoretical basis for these methods, conceptualizing the outputs from expert models as intermediate ``visual thoughts'' that serve as a cache for information transmission. These methods demonstrate that integrating expert systems provides the fine-grained visual information needed for precise spatial grounding and analysis.

\subsubsection{SFT-Based Approaches}
\label{sec:stage1-Approaches-SFT-Based}
SFT is a primary method for teaching LMMs to use external tools or internal visual skills. The process involves fine-tuning a model on datasets that demonstrate the procedural steps of tool use. This includes tool invocation and the integration of tool outputs into a coherent reasoning chain.
\vspace{-5pt}
\paragraph{Learning to Orchestrate Tools}
A primary application of SFT is teaching an LMM to compose and orchestrate a set of external, predefined tools. This approach grounds the model's reasoning in verified, external knowledge sources or specialized visual processors. For example, LLaVA-Plus~\citep{liu2023llavaplus} is fine-tuned on multimodal instruction data, which enables it to manage a ``skill repository'' and learn to activate the appropriate tool based on user intent. A similar strategy is employed by TACO~\citep{ma2024tacolearningmultimodalaction}, which is trained on a large, synthetic dataset of ``Chains-of-Thought-and-Action'' (CoTA) traces. This data explicitly teaches the model to generate reasoning steps, call external tools such as OCR or calculators with the correct syntax, and integrate the returned information. In addition, VTS-V~\citep{bai2025multi} proposes a novel framework where SFT is utilized to supervise the model's generation of multi-step visual reasoning trajectories, enabling dynamic and verifier-guided use of visual information throughout the inference process. In these cases, SFT provides the necessary supervision to translate a high-level plan into a sequence of executable actions, giving the model true procedural competence.

\vspace{-5pt}
\paragraph{Developing Internal Visual Operations.}
Beyond orchestrating external APIs, SFT is crucial for developing a model's capacity for internal visual manipulation. This represents a significant step towards cognitive autonomy, as the skills become part of the model itself. The CogCoM framework~\citep{qi2024cogcom} exemplifies this by teaching a model to perform intrinsic actions like object grounding, regional OCR, or image cropping without relying on external modules. The fine-tuning is performed on a dataset of ``Chain-of-Manipulation'' (CoM) samples. These samples demonstrate not just the action, but also how to generate the necessary parameters for it, such as coordinates for a crop. Similarly, VGR~\citep{wang2025vgr} introduces a novel multimodal large language model that enhances fine-grained visual perception by detecting relevant regions and providing precise answers based on replayed image regions, addressing limitations of relying solely on language space for multimodal chain-of-thought reasoning. Furthermore, UniVG-R1~\citep{bai2025univg} tackles the challenge of universal visual grounding in multi-image scenarios with complex instructions by proposing a reasoning-guided multimodal large language model that utilizes reinforcement learning and a high-quality Chain-of-Thought grounding dataset to enhance its reasoning capabilities.

\vspace{-5pt}
\paragraph{Cultivating Dynamic Perception Exploration.}
A particularly advanced application of SFT is the cultivation of dynamic visual attention as a learned, internal tool. This approach directly addresses the limitations of static perception by enabling the model to actively inspect visual input. This is achieved by fine-tuning the model on datasets with specialized attentional annotations. For instance, Visual CoT~\citep{shao2024visual} uses data with bounding box annotations to teach the model to highlight critical regions during reasoning. In contrast, Instruction-Guided Visual Masking (IVM)~\citep{zheng2024instruction} trains the model to generate masks that hide instruction-irrelevant image areas, effectively sharpening its focus. SFT can also foster more complex cognitive operations, such as learning cross-image comparison from demonstration~\citep{zhang2025cmmcot} or developing a policy for selective visual revisitation~\citep{chung2025don}. Architectures like V*~\citep{vstar} also rely on this principle to train a visual search component. The common thread is that SFT transforms attention from a passive mechanism into an active, controllable skill, 
% with the training data providing direct supervision for the act of looking itself.
with the training data providing direct supervision for visual attention allocation.

\subsubsection{RL-Based Approaches}
\label{sec:stage1-Approaches-RL-Based}
RL advances beyond supervised imitation by enabling models to discover and optimize policies for tool use through environmental interaction and reward feedback. 
This methodology allows models to learn effective strategies for visual reasoning rather than just replicating demonstrated paths.
\vspace{-5pt}
\paragraph{Establishing Foundational RL Principles.} Before complex ``Thinking with Images'' could be realized, foundational work was necessary to establish the viability and core principles of applying RL to visual tasks. Foundational studies like Jigsaw-R1~\citep{wang2025jigsaw} used controlled environments to reveal key principles, such as that RL offers superior generalization over SFT and that complex reasoning patterns appear to be \textit{pre-existing} in models rather than emergent from training. Concurrently, V-Triune~\citep{ma2025one} and VisionReasoner~\citep{liu2025visionreasoner} demonstrated the versatility of the RL approach by creating a unified framework. VisionReasoner, for instance, employs novel multi-object cognitive strategies and tailored rewards to handle diverse perception tasks like detection, chart, and counting within a single shared model. Together, this foundational work proved that LMMs could learn generalizable visual skills via reinforcement learning.
\vspace{-5pt}
\paragraph{Learning Direct Integration Policies.} With the foundational viability of this learning paradigm established, the initial application focused on a direct form of multimodal integration: optimizing policies that produce reasoning steps which are themselves multimodal. This represents a crucial advance beyond purely textual chains of thought by embedding spatial or visual information directly into the model's output stream. To date, this concept has been most prominently realized through policies that learn to associate reasoning with positional information. The GRIT framework~\citep{fan2025gritteachingmllmsthink}, for instance, learns to effectively interleave natural language with explicit bounding box coordinates. Following this, Point-RFT~\citep{ni2025pointrftimprovingmultimodalreasoning} demonstrates how to further optimize these grounded rationales for correctness. The Seg-Zero~\citep{liu2025segzero} framework showcases a decoupled architecture where a model learns a policy for generating positional prompts that guide a separate segmentation model. Although represented as text tokens, these integrated outputs function as a nascent form of visual thought, an intermediate representation that explicitly merges spatial awareness into the reasoning process itself.

\vspace{-5pt}

\paragraph{Exploring Active Tool Orchestration.}
Building upon this, the next logical evolution in ``Thinking with Images'' grants models the ability to actively manipulate the visual input through direct tool orchestration. Initial frameworks centered on the fundamental challenge of tool management. VisTA~\citep{huang2025visualtoolagent}, for example, employs an RL agent to learn tool \textit{selection} strategies, decoupling the agent from the frozen reasoner for policy transfer. This work leads to active perception, where the model learns to dynamically control its visual focus, often through an adaptive ``zoom-in'' capability. Chain-of-Focus~\citep{zhang2025chainoffocusadaptivevisualsearch} creates a ``chain of focus'' by progressively zooming based on visual cues, while ACTIVE-o3~\citep{zhu2025activeo3} trains a sensing policy to propose informative regions that guide a task-execution component. Further advancing this, DeepEyes~\citep{zheng2025deepeyesincentivizingthinkingimages} enables this capability to emerge natively from a model's inherent grounding ability, using an Interleaved Multimodal Chain-of-Thought (iMCoT) without requiring cold-start SFT. Recognizing that models may be reluctant to use these novel operations, Pixel-Reasoner~\citep{pixelreasoner} introduces a curiosity-driven RL scheme to explicitly incentivize this exploration.
The principle of active interaction, however, is not limited to a single skill. VILASR~\citep{wu2025reinforcing}, for instance, proposes a ``drawing to reason in space'' paradigm, enabling models to leverage basic drawing operations to interact visually. This progression from static image viewers to active participants results in the development of frameworks that empower agents to master a diverse set of vision tools in a human-like manner.
Extending beyond a single perceptual skill, OpenThinkIMG~\citep{su2025openthinkimg} introduced the first open-source end-to-end framework to train adaptive policies for \textit{invoking} a diverse set of external vision tools by directly optimizing for task success.

\subsection{Conclusion and Future Frontiers}

\label{sec:stage1-conclusion}
\paragraph{\includegraphics[scale=0.03]{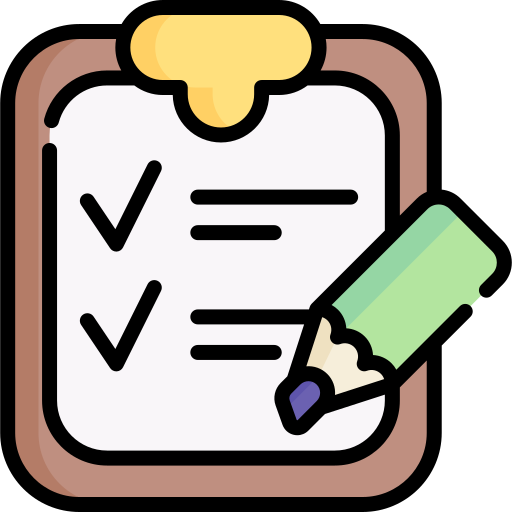} Conclusion.}

The explicit visual tool exploration paradigm is powerful due to its directness and practicality. Its strength lies in allowing models to leverage existing, specialized visual tools to solve specific sub-tasks in a transparent and debuggable manner~\citep{yang2023mmreact, ma2024tacolearningmultimodalaction}. This capability is realized through a methodological spectrum: in-context prompting offers zero-shot flexibility, Supervised Fine-Tuning instills reliable, procedural competence, and Reinforcement Learning enables the discovery of autonomous policies. Despite this versatility, the paradigm is fundamentally constrained by its reliance on a fixed set of predefined tools~\citep{liu2023llavaplus}. A model cannot invent a novel visual operation beyond its given toolbox. This core limitation motivates the move towards the programmatic approaches discussed in the next section, which grants models the ability to construct their own unique visual operations.

\vspace{-5pt}
\paragraph{\includegraphics[scale=0.05]{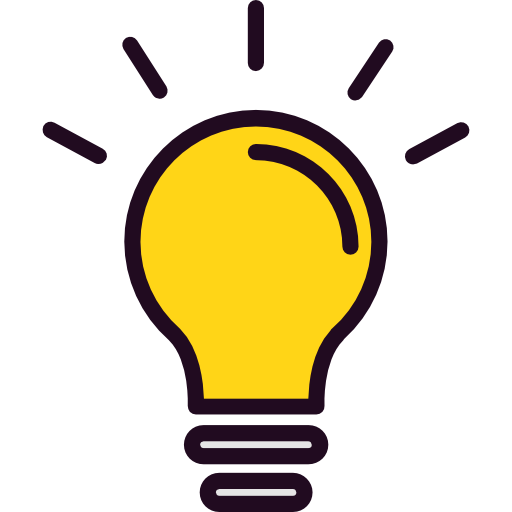} Future Frontiers.}
While the field is advancing, several frontiers remain for tool orchestration. Foundational work will continue to focus on the tools themselves: enhancing their quality to reduce error propagation, expanding their coverage to address long-tail scenarios, and developing standardized interfaces to ensure robust integration. Beyond these practical improvements, however, the most significant frontiers concern the agent's cognitive capabilities.
\begin{itemize}[left=2pt,topsep=1pt,itemsep=2pt, parsep=1pt]
    \item \textbf{Autonomous Policy Learning.} A critical direction is moving beyond simple tool invocation toward autonomous policy learning. This involves training agents that can not only select the right tool but also compose them in novel sequences to solve complex, multi-step problems. The current shift from imitation learning (SFT) to goal-driven discovery (RL) is the first step. Future work will require developing novel methodologies that allow an agent to quickly adapt to new tools and learn generalizable strategies for visual problem-solving.

    \item \textbf{Internalization of Skills.} An even more profound frontier is the internalization of skills, which blurs the line between external tools and internal capabilities. Instead of always relying on an external tool, a future model might develop an innate, fine-tunable skill for ``Thinking with Images''. The current work on using RL to treat visual attention as a controllable ``spotlight'' is an early example~\citep{zhang2025chainoffocusadaptivevisualsearch}. Pushing this frontier means designing architectures and training objectives that encourage models to absorb the functionality of common tools, transforming them from passive orchestrators into agents with a rich set of intrinsic visual abilities.
\end{itemize}

\newcolumntype{C}[1]{>{\centering\arraybackslash}p{#1}}
\newcolumntype{R}[1]{>{\raggedleft\arraybackslash}p{#1}}

\begin{table*}[!t]
\fontsize{7.5}{9.5}\selectfont
\centering
\setlength{\tabcolsep}{1.5mm}
\renewcommand{\arraystretch}{1.1}
\resizebox{\linewidth}{!}{
\begin{tabular}{l >{\raggedright\arraybackslash}p{3.0cm} R{1.4cm} C{0.6cm}C{0.6cm}C{0.6cm} C{0.5cm}C{0.5cm}C{0.5cm}}
\toprule
\multirow{2}{*}[-2pt]{\textbf{Methods}} & \multirow{2}{*}[-2pt]{\textbf{Base Model}} & \multirow{2}{*}[-2pt]{\textbf{Data}} & \multicolumn{3}{c}{\textbf{Operation Categories}} & \multicolumn{3}{c}{\textbf{Scenario}} \\
\cmidrule(lr){4-6} \cmidrule(lr){7-9}
& & & \textbf{VO} & \textbf{LO} & \textbf{CO} & \textbf{P} & \textbf{R} & \textbf{G} \\ 

\midrule
\multicolumn{9}{c}{\textbf{\textit{Prompt-Based Approaches}}} \\
\midrule
VisProg~\citep{gupta2023visual} & GPT-3 & -- & \checkmark & \checkmark & \checkmark & \checkmark & \checkmark  &  \\
ViperGPT~\citep{surismenon2023vipergpt} & GPT-3 & -- & \checkmark & \checkmark & \checkmark & \checkmark & \checkmark & \\
CAD-Assistant~\citep{mallis2024cad} & GPT-4o & -- & \checkmark & \checkmark &  & \checkmark & \checkmark &  \\
MMFactory~\citep{fan2024mmfactory} & LLaVA-7/13B, etc. & -- & \checkmark & \checkmark & \checkmark & \checkmark & \checkmark &  \\
Visual Sketchpad~\citep{hu2024visual} & GPT-4o & -- & \checkmark & \checkmark & \checkmark & \checkmark & \checkmark &  \\
VipAct~\citep{zhang2024vipact} & GPT-4o & -- & \checkmark & \checkmark & \checkmark & \checkmark & \checkmark &  \\
SketchAgent~\citep{vinker2024sketchagent} & Claude3.5-Sonnet & -- & \checkmark & \checkmark &  & \checkmark & \checkmark & \checkmark \\
ReFocus~\citep{fu2025refocus} & GPT-4o & -- & \checkmark & \checkmark &  & \checkmark & \checkmark &  \\
Interactive Sketchpad~\citep{chen2025interactive} & GPT-4o & -- & \checkmark & \checkmark & \checkmark & \checkmark & \checkmark &  \\
\midrule
\multicolumn{9}{c}{\textbf{\textit{Training-Based Approaches}}} \\
\midrule
VPD~\citep{hu2024programdistillation} & PaLI-3 \& PaLI-X & 89.6K & \checkmark & \checkmark & \checkmark &\checkmark & \checkmark & \\
CoSyn~\citep{yang2025scaling} & CLIP \& Mistral-7B & 400K & \checkmark & \checkmark & \checkmark & \checkmark &  &  \\
PROVISION~\citep{zhang2024provision} & xGen-MM4B & 10M & \checkmark & \checkmark &  & \checkmark & \checkmark & \\
Flame VLM~\citep{ge2025advancing} & SigLIP \& DeepSeek-coder & 3.3M+ & \checkmark & \checkmark &  & \checkmark &  &  \\
MathCoder-VL~\citep{wang2025mathcoder} &  InternVL2-8B, etc. & 3M & \checkmark & \checkmark & \checkmark & \checkmark & \checkmark & \checkmark \\
\midrule
Visual-ARFT~$\dagger$~\citep{liu2025VisualARFT} & Qwen2.5-VL-3/7B & 1.2K & \checkmark & \checkmark & \checkmark & \checkmark & \checkmark & \\
% VRAG-RL~\citep{wang2025vrag} & Qwen2.5-VL-3/7B & 70K & \checkmark & \checkmark & \checkmark & \checkmark & \checkmark & \\

\bottomrule
\end{tabular}
}
\vspace{1mm}
\caption{Methods for Programmatic Visual Manipulation. This table classifies approaches into prompt-based and training-based (method marked with $\dagger$ utilizes reinforcement learning, while the rest utilizes supervised fine-tuning). It details the categories of    operations employed: Visual Operations (\textbf{VO}), Logical Operations (\textbf{LO}), and Composite Operations (\textbf{CO}); as well as their application scenarios, spanning Perception (\textbf{P}), Reasoning (\textbf{R}), and Generation (\textbf{G}), similar to Stage 1.}
\vspace{-10pt}
\label{tab:stage2_methods}
\end{table*}

\section{Stage 2: Programmatic Visual Manipulation}
\label{sec:stage2}

While Stage 1 teaches a model to \textit{select} from a fixed toolkit, Stage 2 represents a profound leap in autonomy by teaching it to \textit{create}. Here, the model evolves from a tool orchestrator into a visual programmer. In particular, it learns to generate executable code that specifies tailored visual operations, granting it the flexibility to construct novel analysis pipelines tailored to unique demands.
Table~\ref{tab:stage2_methods} summarizes the main methods in this stage, categorizing them by their composable operations and primary application scenarios.

\subsection{The Programmatic Leap: From Selection to Creation}
\label{sec:stage2-leap}
This programmatic leap from tool selection to code creation unlocks the fundamental advantages of this stage. By moving beyond a finite toolbox to a generative grammar of visual operations, the model gains compositional flexibility, enhanced control, and greater interpretability.

\paragraph{Compositional Flexibility.}
By generating code, the model transcends the constraints of a finite tool menu and gains access to a potentially infinite space of visual operations. It can compose primitive functions to construct novel and highly specific analysis pipelines on the fly~\citep{gupta2023visual, surismenon2023vipergpt}. The core insight is the shift from a finite toolbox to a generative grammar of visual manipulation. This compositional power is critical for tasks with complex, multi-step visual criteria that no single pre-defined tool could address, such as a workflow to ``find all objects whose color matches the average color of a specified region''. This empowers the model with a robust form of generalization, enabling it to construct solutions for entirely new problems from first principles.

\paragraph{Dynamic and Non-Linear Control Flow.}
Orchestrating fixed tools is often a linear process (e.g., call Tool A, then Tool B), where the LMM handles all decision-making between steps. By generating code, the model can embed logical branches (\texttt{if-else}), loops (\texttt{for}), and stateful variables directly into its program~\citep{chen2022program, gao2023pal}. This means the model creates a dynamic program that can adapt its execution path based on intermediate visual findings~\citep{yao2023react, hu2024visual}. For instance, it can first diagnose a visual artifact like image blur and then generate code to apply a deblurring filter before proceeding with its analysis, a capability essential for robust performance in dynamic environments~\citep{liu2025VisualARFT}.

\paragraph{Enhanced Interpretability and Control.}
The generated code serves as a deterministic, and human-readable reasoning trace. This provides a transparent view into the model's problem-solving process, in contrast to the opaque nature of a model's internal thought process or a black-box tool call. Each line of code is an explicit, verifiable step in the model's logic~\citep{gupta2023visual, hu2024programdistillation}. This transparency is not merely for observation; it provides a powerful mechanism for control and collaboration. The code acts as a shared artifact that a human user can inspect, debug, and even refine. This creates a natural entry point for human-AI interaction, enabling an iterative workflow where human expertise can guide and correct the AI's visual analysis, a crucial feature for high-stakes applications~\citep{mallis2024cad, chen2025interactive}.

\subsection{Formulation}
\label{sec:stage2-formulation}
In this paradigm, the LMM's primary action is to generate programmatic instructions \(C\), typically as a script in a language like Python that utilizes computer vision libraries (e.g., OpenCV, Pillow) or domain-specific APIs. The state \(S_t\) includes the history of previously generated code snippets and outputs. The core action at each reasoning step is the generation of a code segment \(C_t\), conditioned on the reasoning history \(S_t\), the initial image \(I\), the original input query \(Q\):
\begin{equation}
C_t \sim P(\cdot | S_t, I, Q; \Theta_{LMM})
\label{eq:think_program_gen_revised}
\end{equation}
These segments are composed into a complete program \(C = (C_1, C_2, \dots, C_N)\). This program is then passed to an external interpreter or execution engine \(\mathcal{E}_{\text{exec}}\), which applies the sequence of operations to an initial image \(I_0=I\):
\begin{equation}
(I_{\text{final}}, O_{\text{prog}}) = \mathcal{E}_{\text{exec}}(C, I_0)
\label{eq:think_program_exec_revised}
\end{equation}

The execution yields a potentially modified image \(I_{\text{final}}\) and any textual or structured output \(O_{\text{prog}}\) generated by the program, such as calculated values or detected features. The LMM uses this information to continue reasoning or to produce the answer. The fundamental distinction from Stage 1 is that the visual operations are no longer selected, but are composed and defined by the LMM through code generation. This allows the model to design and execute tailored, complex algorithms.

\subsection{Categories of Composable Operations}
\label{sec:stage2-categories}
In the programmatic paradigm, the model's vocabulary consists of composable building blocks that it generates as code. As shown in Figure~\ref{fig:evolve}, these operations can be grouped into three main categories:

\begin{itemize}[left=2pt,topsep=2pt,itemsep=1pt]
    \item \textbf{Visual Operations.} These are the fundamental functions that directly interact with image data, acting as the model's ``hands and eyes'' on the visual canvas. This category includes perception-oriented operations like region and object detection (\textit{e.g.}, \texttt{find\_objects}, \texttt{crop\_region}) and feature extraction (\textit{e.g.}, \texttt{get\_color}, \texttt{ocr}). It also includes action-oriented operations such as geometric transformations (\textit{e.g.}, \texttt{rotate}, \texttt{resize}) and, critically, drawing functions that allow the model to create visual aids to guide its own reasoning, such as annotating an image with auxiliary lines or highlighting areas of interest~\citep{bradski2000opencv, chen2025interactive}. These primitives form the essential vocabulary for any visual algorithm.

    \item \textbf{Logical Operations.} These operations leverage the power of the host programming language to provide structure and logic to the visual analysis. This includes state management via variable assignment, conditional execution through \texttt{if/else} blocks, and iterative processing over visual elements using \texttt{for} or \texttt{while} loops~\citep{hu2024programdistillation, ge2025advancing}. This logical scaffolding is the grammar that arranges the visual operations into a coherent and powerful algorithm. It enables the model to perform stateful, multi-step reasoning, transforming a simple sequence of actions into a true computational process.

    \item \textbf{Composite Operations.} This category represents the synergistic combination of visual and logical operations. By weaving visual operations into logical scaffolding, the model can create highly specific and complex algorithms that exhibit emergent behavior. A task like ``find all red circles larger than the average area of all circles'' is a perfect example. It requires a composite algorithm that iterates through objects (loop), checks their properties (conditionals), performs calculations (stateful variables), and ultimately returns a result. This ability to construct tailored algorithms on demand is what distinguishes this stage, moving beyond predefined capabilities to genuine, problem-driven visual computation~\citep{fan2024mmfactory, mallis2024cad}.
\end{itemize}

\subsection{Implementation Approaches}

Similar to the tool orchestration paradigm, programmatic manipulation is enabled by the same three core methodologies: training-free prompting, SFT, and RL methods. This section details how these approaches are specifically adapted to equip models with code-generation capabilities.

\subsubsection{Prompt-Based Approaches}
\label{sec:stage2-Approaches-Prompt-Based}
In this stage, prompt-based approaches instruct an LMM to generate executable code, typically in Python, as a transparent intermediate reasoning step. This method leverages in-context learning to create deterministic and verifiable visual analysis pipelines.

\paragraph{Composing Programs from Instructions} 
This line of work establishes the core paradigm of programmatic visual reasoning by leveraging LLMs to generate executable code that orchestrates visual computations. VisProg~\citep{gupta2023visual} pioneered this approach by utilizing the in-context learning capabilities of GPT-3 to generate Python-like programs from natural language instructions, decomposing complex visual tasks into sequences of modular operations that invoke off-the-shelf computer vision models, image processing functions, and logical operators without requiring task-specific training. Building upon this foundation, ViperGPT~\citep{surismenon2023vipergpt} refined the methodology by employing code-generation models like Codex to produce Python programs that explicitly define reasoning steps through a predefined API, enabling the composition of various pre-trained vision modules as subroutines while leveraging Python's native logic for operations such as counting, sorting, and conditional branching. Both approaches share the fundamental principle that code serves as an interpretable and verifiable intermediate representation for visual reasoning, where the generated program itself becomes the explanation of the reasoning process. This training-free paradigm demonstrates remarkable flexibility and strong zero-shot performance across diverse visual reasoning tasks, establishing code generation as a powerful method for bridging natural language instructions with systematic visual computations through modular, composable operations.

\vspace{-5pt}
\paragraph{Sketching Thoughts onto Canvases.} 
This advanced category extends beyond tool invocation to actively create and manipulate visual content as integral components of the reasoning process, embodying a more refined form of visual thinking.
Visual Sketchpad~\citep{hu2024visual} introduces the concept of a ``visual CoT'' by enabling LMMs to synthesize Python code that generates intermediate visual aids (e.g., auxiliary lines for geometry problems). These visual aids then inform subsequent reasoning steps in an iterative loop of thought, action, and observation.
Complementing this generative approach, ReFocus~\citep{fu2025refocus} focuses on structured image understanding by empowering models to generate Python code for targeted visual editing operations, including drawing bounding boxes and masking irrelevant areas to implement selective attention for tables and charts. SketchAgent~\citep{vinker2024sketchagent} demonstrates language-driven sequential sketch generation through an intuitive sketching language that represents drawings as stroke-by-stroke actions, enabling diverse sketch creation without model training. Interactive Sketchpad~\citep{chen2025interactive} showcases educational applications by generating accurate diagrams through Python code execution, facilitating collaborative problem-solving on shared interactive whiteboards. 
This paradigm establishes programmatic visual manipulation as a powerful reasoning mechanism, enabling models to externalize and refine their visual understanding through iterative code-driven interactions with visual content.

\paragraph{Weaving Expertise into Systems.} 
This most advanced category represents the culmination of programmatic approaches through comprehensive system architectures and deep domain specialization that transcend simple code generation. VipAct~\citep{zhang2024vipact} exemplifies advanced multi-agent integration by orchestrating a collaboration framework with an orchestrator agent, specialized LMM agents, and vision expert models that systematically gather detailed visual evidence across multiple components. MMFactory~\citep{fan2024mmfactory} introduces a universal ``solution search engine'' that employs committee-based multi-agent LLM conversations to generate diverse, executable programmatic solutions while benchmarking performance and computational costs for optimal selection. CAD-Assistant~\citep{mallis2024cad} demonstrates deep domain specialization by creating a comprehensive CAD agent fully integrated with FreeCAD software, enabling iterative interaction with CAD models through specialized tools for parameterization, rendering, and constraint analysis. This paradigm represents the maturation of programmatic visual reasoning into production-ready systems that combine complex architectural design, multi-component orchestration, and domain-specific expertise to tackle real-world applications with practical deployment requirements.

\subsubsection{SFT-Based Approaches}
\label{sec:stage2-Approaches-SFT-Based}
SFT is a crucial method for teaching LMMs to reason programmatically. It fine-tunes a model to either generate code as an intermediate reasoning step or to understand images rendered from code. This approach leverages the structured logic of programs to guide the learning process, pushing the model beyond simple pattern recognition toward more systematic and transparent problem-solving.

\paragraph{Internalizing Logic from Programs.}
This strategy uses SFT to distill multi-step programmatic logic into LMMs. Visual Program Distillation (VPD) ~\citep{hu2024programdistillation}, uses LLMs to generate executable programs that invoke specialized vision modules. The execution traces of verified programs are then converted into natural language rationales, which serve as the data for instruction-tuning the LMM. This distillation process enables the LMM to replicate the detailed programmatic workflow in an efficient, single forward pass. In a similar effort, MathCoder-VL~\citep{wang2025mathcoder} performs an initial ``image-to-code'' SFT stage to align the model's vision encoder with the precise semantics of code (e.g., TikZ, Python) used to render mathematical figures, establishing a crucial link between visual data and its symbolic code representation. PROVISION~\citep{zhang2024provision} presents a programmatic approach employing scene graphs as symbolic representations of images and human-written programs to systematically synthesize vision-centric instruction data, enabling the scalable generation of interpretable, controllable, and accurate multimodal instruction data.

\paragraph{Bootstrapping Datasets with Code.}
This approach leverages code generation to create perfectly aligned, high-quality multimodal data for instruction tuning. Frameworks like CoSyn~\citep{yang2025scaling} prompt a text-only LLM to first generate code (e.g., Python, HTML) that renders a synthetic, text-rich image and then use the same source code as ground truth to create corresponding question-answer pairs and CoT rationales. SFT on such data has proven highly sample-efficient for teaching LMMs to reason over structured images like charts and tables. Likewise, the main training stage of MathCoder-VL~\citep{wang2025mathcoder} fine-tunes the model on a dataset of code-synthesized mathematical figures paired with LLM-generated problems and detailed solutions. The Flame VLM~\citep{ge2025advancing} exemplifies a direct application of this, being SFT-trained on synthesized image-code pairs to translate visual UI designs into executable front-end code.

These approaches collectively demonstrate that SFT is being used to bridge the gap between the symbolic, logical nature of code and programs, and the sub-symbolic, pattern-based nature of LMMs. By training on data derived from or representing code, LMMs are effectively taught to ``think like a programmer'' when approaching visual tasks. The structured, compositional, and often hierarchical nature of code-based SFT data induces correspondingly more structured and compositional reasoning pathways within the LMM. This can be seen as a form of implicit neuro-symbolic learning, where the ``symbolic'' component (the logic inherent in the code or program structure) provides a strong inductive bias that shapes the learning of the ``neural'' component during SFT.

\newcolumntype{C}[1]{>{\centering\arraybackslash}p{#1}}
\newcolumntype{R}[1]{>{\raggedleft\arraybackslash}p{#1}}

\begin{table*}[!t]
\fontsize{7.5}{9.5}\selectfont
\centering
\setlength{\tabcolsep}{1.5mm}
\renewcommand{\arraystretch}{1.1}
\resizebox{\linewidth}{!}{
\begin{tabular}{l >{\raggedright\arraybackslash}p{3.0cm} R{1.4cm} C{0.6cm}C{0.6cm}C{0.6cm} C{0.5cm}C{0.5cm}C{0.5cm}}
\toprule
\multirow{2}{*}[-2pt]{\textbf{Methods}} & \multirow{2}{*}[-2pt]{\textbf{Base Model}} & \multirow{2}{*}[-2pt]{\textbf{Data}} & \multicolumn{3}{c}{\textbf{Thoughts Categories}} & \multicolumn{3}{c}{\textbf{Scenario}} \\
\cmidrule(lr){4-6} \cmidrule(lr){7-9}
& & & \textbf{LR} & \textbf{VS} & \textbf{MD} & \textbf{P} & \textbf{R} & \textbf{G} \\ 

\midrule

\multicolumn{9}{c}{\textbf{\textit{SFT-Based Approaches}}} \\
\midrule
Minigpt-5~\citep{zheng2023minigpt} & MiniGPT-4 \& SD v2.1 & 3.62M & \checkmark & & & \checkmark & & \checkmark \\
GILL~\citep{koh2023generating} & OPT-6.7B, etc. & 3.62M & \checkmark & & & \checkmark& &\checkmark \\
Chameleon~\citep{team2024chameleon} & Chameleon-7/34B & 4.4T+ & \checkmark & & &\checkmark && \checkmark\\
Transfusion~\citep{zhou2024transfusion} & Transformer-7B \& U-Net & 2T & \checkmark & & & \checkmark & &\checkmark \\
Emu2~\citep{sun2024generative} &LLaMA-33B, etc. & 160M+& \checkmark & & & \checkmark& & \checkmark\\
SEED-X~\citep{ge2024seed} & LLaMA-2-13B, etc. & - & \checkmark & & & \checkmark& & \checkmark \\
Next-GPT~\citep{wu2024next} & Vicuna-7B, etc. &  7M & \checkmark & & & \checkmark & & \checkmark \\
Anole~\citep{chern2024anole} & Chameleon-7B & 6K & \checkmark & & & \checkmark & & \checkmark \\
EMU-3~\citep{wang2024emu3} & LLaMA-2 \& MoVQGAN & - & \checkmark & & & \checkmark & & \checkmark  \\
Show-o~\citep{xie2024show} &  Phi-1.5 \& MAGVIT-v2 & 1.05B & \checkmark & & & \checkmark & & \checkmark \\
VILA-U~\citep{wu2024vila} & LLaMA-2-7B, etc. & 15M & \checkmark & & & \checkmark & & \checkmark  \\
Metamorph~\citep{tong2024metamorph} &LLaMA-3.1-8B, etc. & 510M & \checkmark & & & \checkmark & & \checkmark\\
LMFusion~\citep{shi2024llamafusion} & LLaMA-3-8B, etc. & 380M+ & \checkmark & & & \checkmark & & \checkmark \\ 
TokenFlow~\citep{qu2024tokenflow} &  CLIP \& SigLIP ViT  & 760M & \checkmark & & & \checkmark & & \checkmark  \\ 
D-DiT~\citep{li2024dualdiffusionunifiedimage} & SD v3-medium \& DiT & 40M & \checkmark & & & \checkmark & & \checkmark \\
MetaQueries~\citep{pan2025transfer} & LLaVA-ov-0.5B, etc. & 25M & \checkmark & & & \checkmark & & \checkmark \\
UniFork~\citep{li2025unifork} & Qwen2.5-0.5B, etc. & 130M+ & \checkmark & & & \checkmark & & \checkmark \\
BAGEL~\citep{deng2025emerging} & Qwen2.5-7B, etc. & 5.1T & \checkmark & & & \checkmark & & \checkmark \\
Mogao~\citep{liao2025mogao} &  Qwen2.5-3B, etc. & - & \checkmark & & & \checkmark & & \checkmark \\ 
BLIP3-o~\citep{chen2025blip3} & Qwen2.5-VL-7B, etc. & 28M+ & \checkmark & & & \checkmark & & \checkmark \\
Think while Gen.~\citep{chern2025thinking} & Anole-7B & 4.5M & & & \checkmark & \checkmark & \checkmark & \checkmark \\
CoT-VLA~\citep{zhao2025cot} & VILA-U-7B & - & &  & \checkmark & \checkmark & & \checkmark \\
GoT~\citep{fang2025got} & Qwen2.5-VL-3B, etc. & 9.37M & & \checkmark & & \checkmark & & \checkmark \\
Janus~\citep{wu2024janus} & DeepSeek-LLM-1.3B, etc. & 7M+ & \checkmark & & & \checkmark & & \checkmark \\
Janus-Pro~\citep{chen2025janus} & DeepSeek-LLM-7B, etc. & 162M+ & \checkmark & & & \checkmark & & \checkmark \\
Show-o2~\citep{xie2025showo2} & Qwen2.5-1.5/7B, etc. & 75M & \checkmark & & & \checkmark & & \checkmark \\ 
MVoT~\citep{li2025imaginereasoningspacemultimodal} & Anole-7B & 17K & & \checkmark & & \checkmark &  \checkmark & \checkmark \\
\midrule
\multicolumn{9}{c}{\textbf{\textit{RL-Based Approaches}}} \\
\midrule
GoT-R1~\citep{duan2025got} & Janus-Pro-1B/7B & - & & & \checkmark & \checkmark & & \checkmark \\
PARM~\citep{guo2025can} & LLaVA-OV-7B \& Show-o & 400K & & \checkmark & & & & \checkmark \\
T2I-R1~\citep{jiang2025t2i} & Janus-Pro-7B & 6.8K & & \checkmark & & & & \checkmark \\
Visual Planning~\citep{xu2025visualplanningletsthink} & LVM-3B & $\approx$ 1K* & & \checkmark & & & & \checkmark \\
ControlThinker~\citep{han2025controlthinker} & Qwen2.5-VL-7B, etc. & 21.6M & & \checkmark & & & & \checkmark \\

\bottomrule
\end{tabular}%
}
\vspace{1mm}
\caption{Methods for Intrinsic Visual Imagination. This table classifies approaches by their base model and data requirements. It details the categories of generative thoughts employed: Implicit Reasoning via Latent Representations (\textbf{LR}), Explicit Reasoning via Visual Scratchpads (\textbf{VS}), and Interleaved Reasoning via Multimodal Dialogue (\textbf{MD}). It also specifies their primary application scenarios: Perception (\textbf{P}), Reasoning (\textbf{R}), and Generation (\textbf{G}). Data suffixes: * means for each task.}
\label{tab:stage3_methods}
\vspace{-10pt}
\end{table*}

\subsubsection{RL-Based Approaches}
\label{sec:stage2-Approaches-RL-Based}
Reinforcement Learning is a key paradigm for advancing this programmatic approach, as it optimizes a model's ability to generate effective code-based operational sequences by learning directly from execution feedback and task-oriented rewards. This marks a critical shift from imitating static solutions to discovering novel strategies through trial and error, a process that embodies a more dynamic and powerful form of thinking with images.

% RL optimizes a model's ability to generate effective code by learning from execution feedback and rewards. Instead of merely imitating demonstrated solutions, this approach enables the model to discover novel programmatic strategies. This capability is realized through two primary approaches:

% \paragraph{Programming Vision via Feedback}
\paragraph{Visual Agency via Feedback}
RL training enables LMMs to act as autonomous agents, generating code to create tailored visual operations on the fly.
A primary challenge in this paradigm is the vast and sparse space of valid programs; a single error can render a generated script useless. To overcome this, RL-based methods learn from execution feedback.
\mbox{Visual-ARFT}~\citep{liu2025VisualARFT} exemplifies this by employing RL with verifiable rewards. Instead of relying solely on the final task outcome, the model receives crucial intermediate rewards based on whether its generated code executes successfully. This dense reward signal provides a more stable learning gradient, guiding the model to produce functional code and shifting its role from a mere ``tool user'' to a ``tool maker''.

Beyond local image manipulation, this programmatic autonomy extends to foraging for external evidence through agentic web search, a more advanced capability also demonstrated by~\cite{liu2025VisualARFT}. The model learns to seamlessly interleave these two powerful operations, creating a refined, multi-step reasoning workflow. For instance, an agent might first formulate a search query to gather real-time information or contextual knowledge, and then generate Python code to process, analyze, or compare the retrieved information with the original visual input. This interleaved ``search-then-code'' workflow represents a higher level of agentic reasoning, enabling the model to tackle open-world problems that require dynamic information retrieval and precise visual analysis.
\vspace{-5pt}

\subsection{Conclusion and Future Frontiers}
\label{sec:stage2-conclusion}
\paragraph{\includegraphics[scale=0.03]{picture/conclusion.png} Conclusion.}
The programmatic visual manipulation paradigm endows models with compositional flexibility and interpretability. By generating code, a model can construct tailored algorithms from primitive operations, allowing it to tackle a vast range of complex visual problems that are intractable for fixed toolsets~\citep{surismenon2023vipergpt}. The generated program also serves as a transparent and verifiable reasoning trace, which is invaluable for debugging and human-AI collaboration~\citep{wang2025mathcoder}. However, this paradigm's primary limitation is its fundamental dependence on an external execution environment. The reliance on a code interpreter or a search API creates an efficiency bottleneck and a fragile point of failure~\citep{hu2024programdistillation}. More profoundly, a semantic gap persists between the model's internal, latent-space reasoning and the rigid syntax of code. This need to ``outsource'' visual actions motivates the next evolutionary stage: enabling the model to perform these operations intrinsically. This inquiry leads directly to the final stage of our survey, Intrinsic Visual Imagination, where the model generates not an instruction, but the new visual state itself.

\paragraph{\includegraphics[scale=0.05]{picture/bulb.png} Future Frontiers.}
As a critical stepping stone towards intrinsic imagination, advancing the programmatic stage remains a vital research area. Several promising frontiers exist for creating more robust and capable visual programmers.
\begin{itemize}[left=2pt,topsep=1pt,itemsep=2pt, parsep=1pt]
    \item \textbf{Robustness and Self-Correction.} A key direction is developing models that can debug and correct their generated code. This involves training them to interpret error messages from the execution engine or, ambitiously, to visually inspect an incorrect output, diagnose the programmatic flaw, and autonomously rewrite the code. This would create resilient and reliable agents.
    \item \textbf{Unifying Tool Use and Code Generation.} Current models often specialize in either orchestrating tools (Stage 1) or generating code (Stage 2). A significant frontier lies in developing unified agents that take the best of both worlds by combining the advantages of the two stages together. This addresses a fundamental trade-off: the pragmatic efficiency of specialized tools versus the unbounded flexibility of programmatic creation. A truly intelligent multimodal agent should not have to choose. Instead, it should be able to generate a program (Stage 2) that strategically calls pre-existing, high-level tools (Stage 1) as subroutines.

\end{itemize}

\section{Stage 3: Intrinsic Visual Imagination}
\label{sec:stage3}

While Stage 2 teaches a model to \textit{create} programs for external execution, Stage 3 represents the ultimate leap in autonomy by teaching it to \textit{imagine} solutions internally. Here, the model evolves from a visual programmer into an intrinsic visual thinker. It learns to generate new visual states not as instructions to be interpreted, but as a native part of its own thought process, enabling a seamless and closed cognitive loop where it reasons with its own mental imagery.
In Table~\ref{tab:stage3_methods}, we categorize the key approaches of this stage according to the operations they compose and the scenarios they address.

\subsection{The Architectural Leap: From Execution to Imagination}
\label{sec:stage3-leap}
The transition from programmatic manipulation to intrinsic imagination represents an architectural shift. It closes the loop between reasoning and perception by internalizing the ability to create visual content, moving from a model directing external actions to one that performs internal simulations.

\paragraph{Unifying Generation and Reasoning.}
The key innovation of this Stage is the architectural integration of generative and reasoning capabilities within a single, unified model. In prior stages, the model's reasoning (language) and the visual manipulation (execution) are separate processes linked by an API or interpreter. This separation introduces latency and potential information loss. Intrinsic imagination, by contrast, enables a model to generate a new visual state directly from its internal representations, making the act of imagining a native operation within its thought process~\citep{team2024chameleon, sun2024generative}. This unified design removes the bottleneck of external calls and is the foundation for a more fluid and efficient form of multimodal cognition.

\paragraph{From Explicit Instruction to Implicit Simulation.}
Programmatic control requires the model to articulate an explicit, step-by-step procedure in code. Intrinsic imagination allows the model to perform a more holistic and implicit form of simulation. For instance, to predict the outcome of pushing an object, a programmatic model must calculate new coordinates. A model with intrinsic imagination can instead generate an image that directly \textit{depicts} the outcome, implicitly enforcing physical constraints like collision through the visual coherence of the generated image~\citep{xu2025visualplanningletsthink}. This moves the reasoning process from symbolic calculation to perceptual simulation, a mode of thinking that is more robust for complex physical and spatial problems.

\subsection{Formulation}
\label{sec:stage3-formulation}
In this stage, the LMM's core capability is generative visual cognition. Rather than delegating visual tasks, the model directly produces new image states, \(I'_{t}\), as part of its reasoning sequence.

The state \(S_t\) includes the current visual context (which could be the initial image \(I\) or a previously generated image \(I'_{t-1}\)), and the history of textual thoughts and generated images. At a reasoning step \(t\), the LMM's policy determines the modality and content of its next thought:
\begin{equation}
z_{t} \sim P(\cdot | S_t, I, Q; \Theta_{LMM}) \quad \text{where } z_{t} \in \mathcal{T}_{\text{text}} \cup \mathcal{I}_{\text{vis}}
\end{equation}
If the model chooses to generate a visual thought, \(z_{t} = I'_{t}\), this new image is produced directly by the model's internal generative mechanisms (e.g., its decoder in a unified architecture). This generated image \(I'_{t}\) is then fed back as a new perceptual input, becoming part of the state \(S_{t+1}\) for subsequent reasoning. The critical distinction from prior stages is that the generation of the visual step is an \textit{intrinsic capability}, not an outsourced function. This enables a seamless loop between perception, reasoning, and imagination.

\subsection{Categories of Generative Thoughts}
\label{sec:stage3-categories}
Intrinsic visual imagination can be realized through three primary paradigms, each defined by how the model integrates generative acts into its reasoning process. These approaches, also visually represented in Figure~\ref{fig:evolve}, represent different strategies for leveraging internal generative power.

\begin{itemize}[left=2pt,topsep=2pt,itemsep=1pt]
    \item \textbf{Implicit Latent Reasoning.} This paradigm pursues architectural elegance and end-to-end integration. It operates in a unified, efficient, and abstract manner, where visual thinking occurs within the model's abstract feature space by manipulating latent representations or visual tokens that are not necessarily decoded into human-readable images at each step~\citep{team2024chameleon, sun2024generative}. The reasoning is implicit in the transformation of these internal features. The core objective is to create a unified autoregressive architecture that seamlessly predicts the next multimodal element, be it text or visual tokens. This approach prioritizes computational efficiency and a deeply integrated, though less transparent, form of visual thought.

    \item \textbf{Explicit Visual Reasoning.} This paradigm focuses on making the model's reasoning process transparent and interpretable. It achieves this through a transparent, deliberate, and step-by-step process of generating explicit, human-readable images as intermediate steps. This approach extends the textual CoT into a visual domain, creating a ``visual scratchpad'' where the model can externalize its spatial thinking~\citep{guo2025can, li2025imaginereasoningspacemultimodal}. For example, a model might generate an image of a geometry problem with auxiliary lines added, or visualize a future state in a planning task~\citep{xu2025visualplanningletsthink}. This approach makes the model's visual thought process observable, analyzable, and more aligned with human problem-solving methods like sketching.

    \item \textbf{Interleaved Multimodal Reasoning.} This paradigm represents a dynamic interplay between textual and visual thought, where reasoning is constructed through a dynamic, synergistic, and conversational sequence of generated text and images~\citep{duan2025got}. The model's thought process becomes a multimodal dialogue with itself: it might generate text to form a hypothesis, create an image to test it, and then generate more text to reflect on the visual evidence~\citep{chern2025thinking}. This synergistic approach combines the logical structure of language with the grounded intuition of vision, enabling a powerful reasoning process for complex tasks like robotic planning and creative generation~\citep{zhao2025cot}.
\end{itemize}

\subsection{Implementation Approaches}

Unlike prior stages centered on delegation, intrinsic visual imagination repurposes the model's own generative architecture as a reasoning tool. This is achieved primarily through SFT and RL, which shift the learning objective from directing external tasks to performing internal simulation.

\subsubsection{SFT-Based Approaches}
\label{sec:stage3-Approaches-SFT-Based}
SFT is the foundational technique for teaching a model the grammar of multimodal thought. It works by showing the model explicit examples of how to generate interleaved text and images, create intermediate visual representations, or perform visual edits as part of a reasoning chain.

\paragraph{Unifying Modalities through Autoregression.}
A significant push in this area is the development of unified models where SFT is crucial for balancing understanding and generation. Models like Chameleon~\citep{team2024chameleon} and Emu2~\citep{sun2024generative} are trained on vast quantities of interleaved mixed-modal data using a unified autoregressive objective to predict the next multimodal element. SFT is then applied via specialized alignment recipes or instruction tuning to hone performance on specific tasks. A diverse range of SFT strategies has emerged to manage the understanding-generation trade-off~\citep{xie2024large,he2024llms,chen2024next}.  Some models adopt a sequential SFT approach, first training on understanding tasks and subsequently on generation, as seen with BLIP-3o~\citep{chen2025blip3}, which uses a curated instruction-tuning dataset for the generation phase. Others pursue efficiency by fusing pre-trained frozen components; GILL~\citep{koh2023generating}, for example, applies SFT only to lightweight mapping layers that connect a frozen LLM with image encoders/decoders. Further techniques include adjusting data ratios~\citep{chen2025janus}, creating special visual tokens~\citep{tong2024metamorph, zheng2023minigpt}, and freezing text-specific modules to preserve language capabilities while training new visual components~\citep{shi2024llamafusion}. Finally, models like BAGEL~\citep{deng2025emerging} demonstrate that SFT on trillions of tokens from highly diverse, interleaved web data can unlock emergent capabilities like free-form image manipulation. These varied strategies highlight a core design challenge: carefully orchestrating SFT to create synergy, rather than destructive interference, between a model's dual capabilities.

\paragraph{Externalizing Thoughts as Images.}
Another direction uses SFT to enable models to generate visual thoughts as part of a transparent reasoning process. This approach moves beyond text-only chains of thought by externalizing the model's visual and spatial reasoning. For instance, GoT~\citep{fang2025got} uses SFT to train a model to first produce an explicit language plan before generating an image that respects prescribed relationships. MVoT~\citep{li2025imaginereasoningspacemultimodal} fine-tunes models to think visually by generating image visualizations of their reasoning traces. Similarly, CoT-VLA~\citep{zhao2025cot} prompts models to predict future image frames as visual subgoals before generating robotic actions. Others apply reasoning directly to the image creation pipeline. A more advanced paradigm, seen in Thinking with Generated Images~\citep{chern2025thinking}, involves SFT on curated datasets of reasoning chains to enable models to spontaneously generate visual thinking steps like subgoals and self-critiques. This method of using SFT to create explicit multimodal reasoning chains that function as a visual scratchpad marks a shift towards more interpretable and robust AI.

\paragraph{Fusing Reasoning and Generation.}
A clear trend is the convergence of techniques for generation and reasoning, with SFT acting as the unifying mechanism. Reasoning frameworks are increasingly applied to refine generation. For example, T2I-R1~\citep{jiang2025t2i} integrates a bi-level reasoning process into the text-to-image pipeline, using SFT and RL to improve output quality. Conversely, advanced generative mechanisms like diffusion models~\citep{croitoru2023diffusion} are being integrated into reasoning frameworks. BLIP3-o~\citep{chen2025blip3}, for instance, uses SFT to train a diffusion transformer that generates rich image features as a basis for high-quality image synthesis. This interplay shows that robust generation demands advanced reasoning, while powerful reasoning is amplified by the ability to generate intermediate steps. SFT facilitates this powerful integration, blurring the lines between generative models and reasoning models and paving the way for more holistic AI systems.

\subsubsection{RL-Based Approaches}
\label{sec:stage3-Approaches-RL-Based}
While SFT teaches a model to imitate demonstrated reasoning patterns, RL empowers it to autonomously discover effective, multi-step generative strategies to achieve a goal.

\paragraph{Simulating Futures with Vision.}
One of the most advanced forms of thinking with images is \textit{imagination}, in which a model not only perceive or interact with existing visual information but also generates entirely new visual content as intrinsic steps within its reasoning process, representing a move from visual analysis to visual simulation and planning. The clearest embodiment of this concept is found in Visual Planning~\citep{xu2025visualplanningletsthink}, which proposes a framework where models reason and plan purely through sequences of generated images, completely independent of text. This approach is trained with a novel Visual Planning via Reinforcement Learning framework. It demonstrates that a model can successfully navigate complex environments by imagining the sequence of future visual states, setting a benchmark for true visual-native reasoning.

\paragraph{Orchestrating Imagination with Text.} While pure visual planning remains a nascent field, a more common approach bridges textual reasoning with visual imagination. These methods can be understood as a form of ``planned imagination'', where the model first creates a high-level conceptual plan or script in text, and then executes this plan by generating the corresponding visual scene. This is often realized through a bi-level CoT process. For instance, \mbox{GoT-R1}~\citep{duan2025got} applies RL to the Generation CoT framework, enabling models to autonomously discover effective semantic-spatial reasoning plans before generating the image. Similarly, \mbox{T2I-R1}~\citep{jiang2025t2i} introduces a framework that explicitly decouples this process into a semantic-level CoT for high-level planning and a token-level CoT for the patch-by-patch pixel generation, jointly optimizing both with RL. These works demonstrate how the reasoning capabilities of LMMs can be effectively transferred to the visual generation domain, transforming a simple prompt into a well-reasoned visual output.
To bridge the semantic gap in controllable image generation, ControlThinker~\citep{han2025controlthinker} employs a fine-tuned LMM as a ``semantic interpreter''. This MLLM reasons about the low-level control signal to generate an enriched, semantically dense text prompt, which then guides an unmodified generator to create images with superior semantic consistency and quality.

\paragraph{Refining Thoughts via Reflection.} Effectively training these complex, imaginative processes requires advanced RL methodologies that go beyond simple correctness rewards. A key challenge is evaluating the quality of a generated image, which serves as an ``imagined thought''. To address the reward modeling problem, recent work~\citep{guo2025can} introduces PARM and PARM++, novel reward models designed specifically for autoregressive generation. They adaptively assess the generation process step-by-step for clarity and potential, with PARM++ even incorporating a reflection mechanism for self-correction, providing a more in-depth learning signal.

\subsection{Conclusion and Future Frontiers}
\label{sec:stage3-conclusion}
\paragraph{\includegraphics[scale=0.03]{picture/conclusion.png} Conclusion.}
The emergence of intrinsic visual imagination represents the culmination of the ``Thinking with Images'' paradigm, marking a transition towards true cognitive autonomy. Instead of merely processing visual data, models are learning to use vision as an internal medium for thought. The three distinct paradigms discussed each advance this capability in a complementary way. Unified models provide a versatile and scalable foundation, efficiently balancing understanding and generation. Explicit reasoning via visual scratchpads introduces crucial transparency, making the model's thought process interpretable and verifiable, which is invaluable for complex spatial problems. Finally, interleaved multimodal dialogue achieves a powerful synergy, combining the logical structure of language with the grounded intuition of vision in a dynamic reasoning loop. Collectively, these methods chart a clear trajectory for AI, moving it beyond merely consuming visual content to genuinely thinking with it. 
\vspace{-10pt}
\paragraph{\includegraphics[scale=0.05]{picture/bulb.png} Future Frontiers.}
While this progress is transformative, the journey into artificial imagination has only just begun, opening up several critical frontiers for future research. The path for intrinsic visual imagination involves not only refining existing methods but also exploring fundamental questions about the nature of visual thought. The most promising frontiers aim to build more capable, efficient, and collaborative imaginative agents.

\begin{itemize}[left=2pt,topsep=1pt,itemsep=2pt, parsep=1pt]
    \item \textbf{Beyond Photorealistic Imagination.} A crucial frontier involves questioning the necessity of generating photorealistic images for every thought. Human cognition often relies on abstractions and sketches~\citep{weng2025caption,lure,zhou2024aligning}, not perfect mental photographs. Future research should explore a spectrum of internal visual representations beyond raw pixels~\citep{jiang2025vlm}, such as sparse feature maps, semantic masks, or 3D scene graphs. Such abstract representations could offer significant advantages in computational efficiency and might lead to more generalizable and interpretable reasoning.

    \item \textbf{Learning Dynamic World Models.} For imagination to be useful, it must be grounded in a coherent understanding of the world. A major research direction is the development of robust internal world models. This requires training models to simulate not just static scenes but also the dynamics, physics, and causal relationships within an environment~\citep{hao2023reasoning}. Learning these implicit rules will enable more consistent and physically plausible visual simulations, which are essential for robust long-horizon planning and problem-solving.

    \item \textbf{Human-AI Collaborative Imagination.} As AI develops more advanced  imaginative capabilities, a transformative frontier will be human-AI co-imagination. This involves designing interfaces and interaction protocols that allow humans to guide, refine, and co-create with an AI's visual thought process~\citep{chen2025interactive}. Such collaboration could unlock unprecedented creativity in fields from scientific discovery to artistic expression. This also brings forth critical ethical considerations regarding the responsible deployment of AI that can fabricate realistic visual content, necessitating robust safeguards and transparency.
\end{itemize}

\section{Evaluations \& Frameworks for Thinking with Images}
\label{sec:evaluation}

\begin{figure*}[t]
    \centering
    \includegraphics[width=1\textwidth]{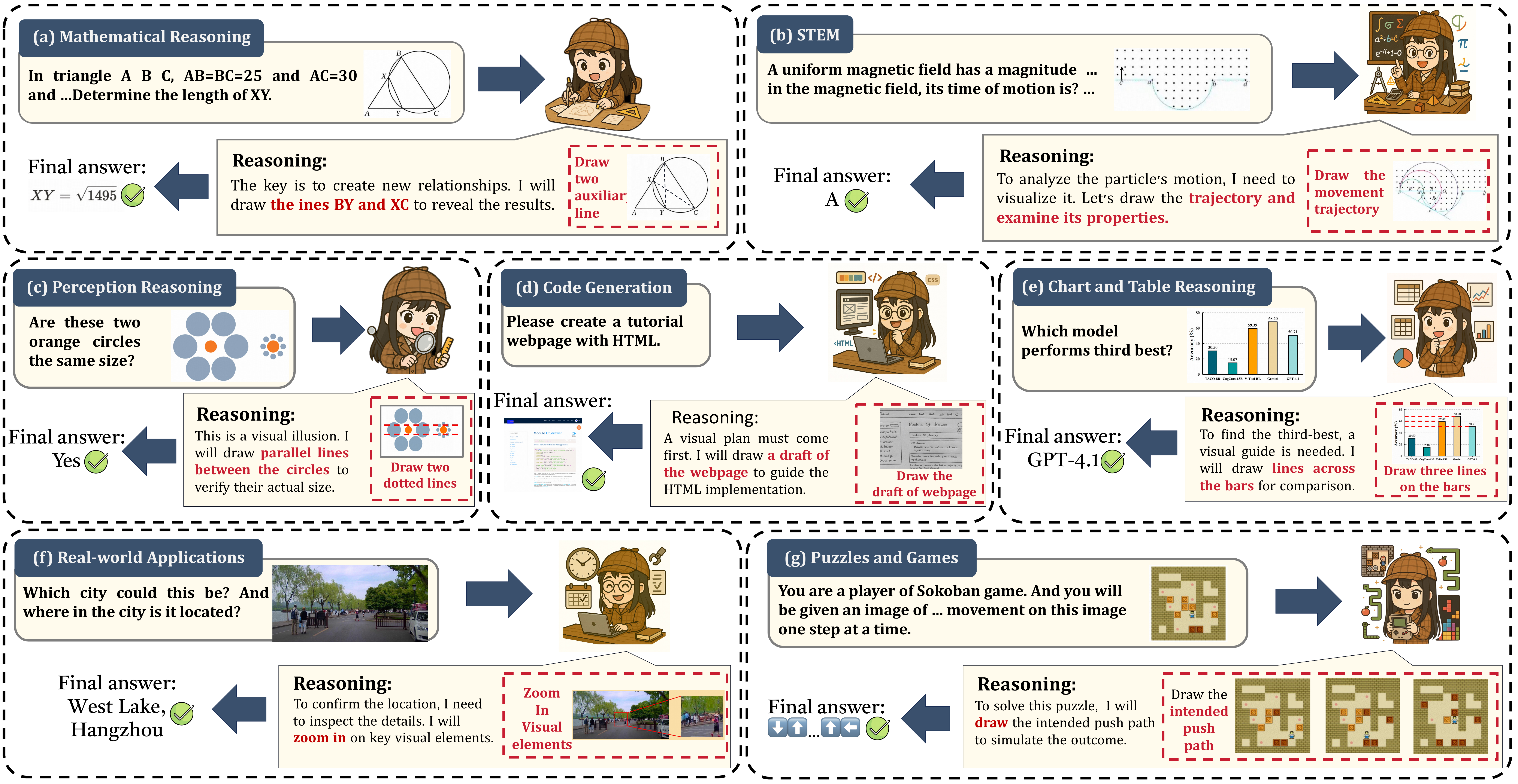}
    \caption{Demonstration of the ``Thinking with Images'' paradigm across seven key evaluation domains. The figure showcases applications ranging from (a) Mathematical Reasoning and (b) STEM to (c) Perception Reasoning, (d) Code Generation, (e) Chart and Table Reasoning, (f) Real-world Applications, and (g) Puzzles and Games.}
    \label{fig:benchmark_examples} 
    \vspace{-10pt}
\end{figure*}
The development and assessment of models capable of ``Thinking with Images'' rely on robust benchmarks and specialized implementation frameworks. This section provides a comprehensive overview of this critical infrastructure. We first review the key benchmarks designed to assess these advanced visual reasoning capabilities. Following this, we describe the primary implementation frameworks that enable models to execute these complex reasoning processes.

\subsection{Evaluations Benchmarks}
\label{ssec:bench}
The benchmarks for ``Thinking with Images'' measure a model's ability to perform complex tasks that extend beyond static visual perception. As illustrated by the diverse examples in Figure~\ref{fig:benchmark_examples}, these tasks span multiple reasoning domains, from mathematical problem-solving to real-world applications. To provide a structured overview of this critical landscape, we organize these benchmarks into seven primary categories based on their core objectives.

\vspace{-5pt}
\paragraph{Mathematical Reasoning.}
Multimodal mathematical problems are designed to evaluate the capabilities of LMMs in complex visual perception, multi-step reasoning, and the integration of mathematical knowledge. Within this domain, geometry problems represent a key challenge in this domain~\citep{cao-geoqa+-coling-2022, chen-unigeo-emnlp-2022, fu-Trustgeogen-arxib-2025, zhang-geoeval-acl-2024}, as they require LMMs to perceive fundamental visual elements and perform multi-step logical reasoning, thereby serving as an effective benchmark for evaluating both the visual and linguistic reasoning capabilities of LMMs. To further assess the reasoning capabilities of LMMs across a broader range of multimodal mathematical scenarios beyond geometry, recent studies have expanded the evaluation scope~\citep{lu-mathvista-iclr-2024, sun-mmmath-emnlp-2024, wang-mvmath-arxiv-2025, sun-mathglance-arxiv-2025} to encompass diverse mathematical tasks, including algebra, analytic geometry, and arithmetic. More 
recently, with the emergence of increasingly powerful multimodal reasoning models~\citep{seedteam-seed1.5vl-arxiv-2025, gemini-team-gemini2.5-report-2025}, an increasing number of expert- and competition-level benchmarks~\citep{wang-mathvision-neurlps-2024, he-olympiadbench-acl-2024, Gupta-Polymath-arxiv-2025} have been introduced to better guide model development and evaluate their reasoning capabilities. Building on these efforts, some studies have been proposed to evaluate multimodal mathematical reasoning from model robustness to diagrammatic variations~\citep{zhang-mathverse-eccv-2024, zou-dynamath-iclr-2025}, reasoning with minimal or no textual context~\citep{Kamoi-VisOnlyQA-arxiv-2024, xiao-logicvista-arxiv-2024}, and generating structured outputs such as visual diagrams or natural language explanations~\citep{li-visiomath-arxiv-2025, Jaewoo-MATHEXPLAIN-2025-arxiv}. The core challenge of multimodal mathematical reasoning lies in performing complex mathematical reasoning in the language space while simultaneously constructing intermediate visual content that is not explicitly provided in the input, such as drawing auxiliary lines or plotting function graphs. Therefore, success on multimodal mathematical benchmarks inherently requires dynamic reasoning on visual content, making them a crucial testbed for evaluating the ``Thinking with Images'' capabilities of LMMs.
\vspace{-5pt}
\paragraph{STEM.} 
STEM (Science, Technology, Engineering, and Mathematics) benchmarks are designed to evaluate the reasoning capabilities of LMMs across scientific and technical domains. In this domain, early efforts focus on broad science education benchmarks, aiming to assess reasoning and knowledge across various subjects at the K–12 level~\citep{lu-scienceqa-nips-2022, hao-emma-arxiv-2025, zhou-mdk12bench-arxiv-2025, ye-mmscibench-arxiv-2025}. As the reasoning abilities of LMMs have rapidly advanced, subsequent works extend evaluations to university-level tasks that require deeper subject knowledge and complex reasoning capabilities~\citep{yue-mmmu-cvpr-2024, xia2024mmie, wang-physunibench-arxiv-2025}, and more recently, to professional-level benchmarks targeting expert reasoning in highly specialized domains~\citep{yue-mmmupro-arxiv-2024, zhou-sfe-arxiv-2024}.
Unlike the constructive but often static reasoning observed in mathematical tasks, STEM problems demand dynamic reasoning over temporal and causal phenomena, including physical motion, chemical reactions, and mechanical operations. This shift enhances the ``Thinking with Images'' paradigm by redefining the image as a dynamic representation of real-world processes. As a result, these benchmarks provide a testbed for evaluating a model’s understanding of physical laws and, more broadly, the sophistication of its emerging internal world model.
\vspace{-5pt}
\paragraph{Puzzles and Games.} 
Benchmarks for puzzles and games are designed to evaluate the capacity of LMMs for strategic planning and reasoning in dynamic, rule-governed environments. Early benchmarks introduced a range of tasks, from visual puzzles assessing logical and spatial reasoning~\citep{zhang-ingvp-arxiv-2024, ren-vgrp-arxiv-2025, zhang-PuzzleBench-arxiv-2025, wu-vsp-arxiv-2024} to interactive games that require understanding of game mechanics and state transitions~\citep{wang-palygame-iclr-2025, tong-code2logic-arxiv-2025, Paglieri-BALROG-arxiv-2024}. More recent benchmarks have increased in complexity, incorporating dynamic video-based contexts that test physical and commonsense reasoning~\citep{zhang-videogamebench-arxiv-2025, cao-PhysGame-arxiv-2024, li-vcbench-arxiv-2024}, as well as unstructured, competition-style puzzles demanding creative and open-ended problem solving~\citep{wang-enigmaeval-arxiv-2024, li-puzzleworld-arxiv-2025}. Distinct from reasoning in STEM domains, the core challenge of games lies in long-horizon, prospective reasoning. This represents a form of ``Thinking with Images'' in which a model must visually simulate complex action sequences on an interactive game board in order to devise an optimal plan. The scalability and structured nature of these environments also makes them suitable for both training and evaluating the planning and reasoning capabilities of LMMs.

\vspace{-5pt}
\paragraph{Perception Reasoning.} 
Perception and reasoning benchmarks evaluate the visual capabilities of LMMs, encompassing the accurate interpretation of fine-grained visual elements and multi-step visual reasoning. One line of these tasks focuses on perceptual acuity, using high-resolution and detail-rich imagery to assess fine-grained visual recognition~\citep{vstar, huang-blink-2024-eccv, wang2025divide}. Another line targets complex cognitive functions, including multi-step reasoning~\citep{bi2025verify}, object hallucination detection~\citep{li-etal-2023-evaluating}, and cognitive evaluations~\citep{song2024cognitive, gupta-risebench-2024-arxiv, huang2023vbenchcomprehensivebenchmarksuite}. These benchmarks assess whether a model can move beyond superficial pattern recognition to develop a deeper, structured understanding of visual scenes that supports robust reasoning. This capability establishes the perceptual and cognitive foundation necessary for the ``Thinking with Images'' paradigm.
\vspace{-5pt}
\paragraph{Code Generation.} 
Code generation from visual or textual inputs represents a key application domain of the ``Thinking with Images'' paradigm, requiring LMMs to generate executable programming logic based on complex visual content and corresponding instructions. Benchmarks in this area evaluate this ability across diverse contexts, such as interpreting mathematical plots and general programming-related diagrams~\citep{li-mmcode-arxiv-2024, wu-plot2code-naacl-2025}. One significant application is front-end development, where models are tasked with converting visual designs, ranging from hand-drawn sketches to polished UI screenshots, into functional webpage code~\citep{si-design2code-naacl-2025, li-Sketch2Code-naacl-2025, Lauren-websight-arxiv-2024}. This visual-to-code process exemplifies a distinctive form of ``Thinking with Images.'' It requires the model to act as a visual decompiler, parsing spatial layouts and design hierarchies to reconstruct a visual artifact into its underlying structural logic. The ability to convert visual intent into precise, executable code represents a practically valuable application of advanced reasoning LMMs.
\vspace{-5pt}
\paragraph{Chart and Table Reasoning.} 
Chart and table reasoning benchmarks evaluate the ability of LMMs to interpret and analyze structured visualizations. These benchmarks assess data literacy through tasks such as chart type classification, data extraction, and question answering on individual charts and tables~\citep{xia-chartx-arxiv-2024, xu-chartbench-arxiv-2023, wang-chartxiv-nips-2024, wu-TableBench-aaai-2025}. Tasks also require robust reasoning over OCR-derived text~\citep{huang-ocrr-arxiv-2025}. More advanced evaluations test deeper comprehension through multi-table and multi-chart comparisons, data summarization, and code generation for chart reproduction~\citep{zhu-MultiChartQA-naacl-2025, tang2025chartmuseum, li-M3SciQA-emnlp-2024, shi-chartmimic-arxiv-2024}. The core challenge in this setting is not limited to data parsing. It involves understanding the visual relationships embedded in the structure of a chart. Meeting this challenge requires a more advanced form of ``Thinking with Images'' in which a model internally transforms the visualization. For example, it may highlight relevant bars or trace trend lines to clarify relationships before performing the logical or computational steps required for analysis.

\vspace{-5pt}
\paragraph{Real-world Applications.} 
Real-world application benchmarks evaluate the ability of LMMs to handle diverse, authentic tasks that reflect practical scenarios from daily-life contexts. Work involves comprehensive, multi-task benchmarks that use authentic materials like examination questions to assess broad competency across a wide range of subjects~\citep{zhu-multi-arxiv-2024, zhang-m3exam-nips-2023, zhang2025mmerealworld, chen-megabench-arxiv-2024}. In parallel, specialized benchmarks probe performance on specific daily-life skills, such as temporal reasoning with clocks and calendars, or visual numerical reasoning and counting~\citep{weng-VisNumBench-arxiv-2025}. The core challenge of these benchmarks is to apply reasoning to noisy, unstructured visual data by leveraging broad commonsense and world knowledge. This serves as the ultimate test for the ``Thinking with Images'' paradigm, assessing whether the LMMs' visual reasoning capabilities can generalize from controlled settings to practical, real-world utility.

\subsection{Implementation Frameworks}
\label{ssec:framework}
In this section, we introduce the frameworks for implementing ``Thinking with Images'' methods through prompt-based, SFT-based, and RL-based approaches mentioned above.

\paragraph{Prompt-based Methods.}
Prompt-based methods provide an accessible and lightweight approach for enabling models to ``Thinking with Images''. General frameworks such as LangChain~\citep{langchain} offer extensive tool integration and support for multimodal inputs through content blocks. CrewAI~\citep{crewai} and AutoGPT~\citep{AutoGPT}, by contrast, emphasize autonomous agents coordination to address complex reasoning tasks. In retrieval-augmented scenarios, LlamaIndex~\citep{llamaindex} excels in data indexing with metadata support for multimodal content, while Haystack~\citep{Haystack} focuses on production-ready pipeline architectures tailored to multimodal systems. Unlike the previously discussed systems, FlowiseAI’s~\citep{FlowiseAI} and AutoChain’s~\citep{AutoChain} frameworks prioritize ease of use, offering user-friendly tools for quick prototyping and development. However, these frameworks still fall short in enabling LMMs to effectively utilize visual tools in their reasoning processes. To address this gap, recent work has adapted foundational infrastructures to support interactive visual thinking. For instance, MM-REACT integrates with systems like LangChain to orchestrate vision experts, whereas VisProg~\citep{gupta2023visual} and ViperGPT~\citep{surismenon2023vipergpt} introduce self-contained programmatic frameworks where the generated code itself serves as the reasoning plan. Building upon this code-generation paradigm, Visual Sketchpad~\citep{hu2024visual} and ReFocus~\citep{fu2025refocus} utilize modern agentic frameworks such as AutoGen, allowing models to generate or modify images within an interactive reasoning manner.

\paragraph{SFT-based Methods.}
Supervised fine-tuning provides an efficient way for enabling models to initialize and learn ``Thinking with Images'' patterns, and several frameworks have emerged to simplify and scale this process. Comprehensive platforms like LLaMAFactory~\citep{LlamFactory} and Axolotl~\citep{Axolotl} emphasize flexible SFT workflows and provide unified SFT interfaces for more than 100 LLMs and LMMs. To alleviating computational effciency, Unsloth~\citep{Unsloth} focuses on accelerating SFT training and reducing memory consumption through custom GPU kernels, achieving up to 60\% memory reduction and 30x speedup improvements particularly beneficial for resource-constrained multimodal fine-tuning scenarios. For enhanced fine-grained control, Torchtune~\citep{Torchtune} offers precise SFT control through composable building blocks and native PyTorch implementations, enabling custom training loops and specialized data processing pipelines. For large-scale distributed training scenarios, Megatron-LM~\citep{shoeybi2019megatron} provides robust tensor and pipeline parallelism capabilities specifically designed for massive transformer models, enabling efficient scaling across hundreds of GPUs. Additionally, SWIFT~\citep{zhao2024swiftascalablelightweightinfrastructure} offers a scalable and lightweight infrastructure that streamlines the SFT process with minimal computational overhead while maintaining training effectiveness. In contrast to these general training frameworks, NVIDIA's NeMo AutoModel~\citep{NeMoAutoModel} provides enterprise-grade SFT capabilities with native optimization for NVIDIA hardware and comprehensive support for various parallelism strategies and precision configurations. Building upon these general-purpose training frameworks, methods for ``Thinking with Images'' introduce specialized adaptations, primarily by curating highly structured SFT data. This data explicitly models multi-step reasoning through sequences of tool invocations, programmatic execution traces, or generative visual thought processes, thereby teaching the model complex cognitive skills beyond simple instruction following.

\paragraph{RL-based Methods.}
Reinforcement learning offers a powerful mechanism for enabling LMMs to acquire and internalize ``Thinking with Images'' behaviors through interaction with environments. To support this capability at scale, several frameworks have been developed to support scalable and customizable RL workflows on various LMMs. VeRL~\citep{verl_software_2024} achieves scalable RLHF solutions by integrating FSDP and Megatron-LM backends to optimize throughput, and has gained widespread adoption in academic and industry communities. To address high-performance distributed training requirements, OpenRLHF~\citep{hu2024openrlhf} employs a ray-based architecture augmented with vLLM acceleration and advanced GPU-sharing mechanisms, significantly improving training speed over alternative methods. For efficient scaling across large GPU clusters, ROLL~\citep{roll2025alibaba} utilizes multi-role distributed architectures with ray and integrates megatron-core and vLLM to acclerate model training and inference. In contrast to these training optimization frameworks, TRL~\citep{vonwerra2020trl} integrates seamlessly with the Hugging Face ecosystem to facilitate rapid development. To enhance the flexibility of frameworks, RL4LMs~\citep{ramamurthy2022reinforcement} and TRLX~\citep{havrilla-etal-2023-trlx} provide modular, customizable components specifically designed for alignment research and novel algorithm development. Despite the flexibility and scalability of these frameworks, they remain limited in supporting LMMs to ``Thinking with Images'' through interactive learning. To address this limitation, recent approaches have extended general-purpose RL methods with specialized mechanisms tailored for visual reasoning. For example, OpenThinkIMG~\citep{su2025openthinkimg} introduces an end-to-end framework that learns tool orchestration policies, integrating disparate vision tools within a unified optimization loop. In contrast, VILASR~\citep{wu2025reinforcing} proposes a novel ``drawing'' action space that enables direct image manipulation for spatial reasoning tasks. Building upon the VeRL, VisionReasoner~\citep{liu2025visionreasoner} implements a unified multi-task framework that learns to route queries to specialized perception heads, jointly optimizing both routing and execution policies within a single RL iteration.

\subsection{Conclusion and Future Frontiers}

\paragraph{\includegraphics[scale=0.03]{picture/conclusion.png} Conclusion.}
The development of evaluation and implementation frameworks for multimodal reasoning is progressively advancing the concept of ``Thinking with Images'' from a theoretical paradigm to an empirical research domain. On the evaluation front, emerging benchmarks demonstrate the advantages of LMMs with ``Thinking with Images'' capabilities over those employing text-centric reasoning paragdigm. These benchmarks are also evolving from single-turn evaluations to multi-step, multi-task reasoning, mirroring the shift in academia and industry toward multi-round reinforcement learning, tool-augmented thinking, and multi-task reasoning. To enable ``thinking with images'' in LMMs, existing implementation frameworks based on prompting, SFT, and RL provide essential support with only minor adjustments. As LMMs with this reasoning paradigm demonstrate significant performance improvements, these general-purpose frameworks will offer more streamlined and customized support. Overall, these foundations of evaluation and implementation transform into assessable, reproducible, and scalable research practices, supporting the development of more robust and capable multimodal reasoning models.
\vspace{-10pt}
\paragraph{\includegraphics[scale=0.05]{picture/bulb.png} Future Frontiers.}
The established evaluation and implementation frameworks provide a solid foundation for this paradigm. To achieve further advancement, several promising directions in evaluation and implementation frameworks exist for developing more powerful LMMs.

\begin{itemize}[left=2pt,topsep=1pt,itemsep=2pt, parsep=1pt]
    \item \textbf{Benchmarks for Visual Manipulation and Construction.} While the number of multimodal reasoning benchmarks is growing, few require the multi-step reasoning and imagination necessary to arrive at a final answer. This gap limits the ability to evaluate models truly capable of ``Thinking with Images''. Therefore, future benchmarks could include tasks that require the manipulation and imagination of complex visual elements in the given images. For example, a model might be required to draw an auxiliary line on a geometry diagram or sketch a navigational path through a maze. Furthermore, evaluating a model's capacity for complex, real-world multimodal reasoning is of great significance. Such evaluations would prompt models to generate a sequence of intermediate visual states, in guiding the development of world models.
 
    \item \textbf{Enhancing Framework Efficiency and Scalability.} Future work could focus on making implementation frameworks more efficient and scalable. Prompting frameworks will evolve to streamline the integration of visual tools, enabling lightweight commands for generation or editing to be embedded within the reasoning process. In parallel, SFT frameworks should become more efficient at training on diverse datasets, helping models internalize behaviors like tool use and programmatic logic without prohibitive costs. Finally, advancing RL frameworks requires a dual focus on rollout and training efficiency: the policy rollout should be supported by robust, low-latency execution engines, while the policy update could leverage more data-efficient supervision signals from intermediate steps like code execution feedback, moving beyond sparse, outcome-based rewards.

\end{itemize}

\section{Applications}
\label{sec:application}

LMMs capable of ``Thinking with Images'' transform a diverse range of applications, enabling a more in-depth understanding of visual information and more intuitive human-machine interaction. This section explores key domains where these advanced capabilities show significant promise.

\subsection{Interactive Systems and User Interfaces}
\label{ssec:app_interactive_systems}

LMMs are increasingly applied to Graphical User Interfaces (GUIs), where agents must perceive on-screen content and plan actions in a human-like manner~\citep{cheng2024seeclick, zheng2024gpt, wang2024mobile, qin2025ui, xie2024osworld, wu2024copilot, gou2024navigating}. Recent work explores how models can ``Thinking with Images'' to understand interfaces and perform interaction tasks. Foundation models like CogAgent~\citep{Hong2024CogAgent} utilize high-resolution vision encoders to recognize fine-grained UI elements directly from pixels. This approach achieves state-of-the-art results on GUI benchmarks and can outperform agents that rely on extracted HTML, demonstrating the power of direct visual perception. Similarly, MobileLMM~\citep{liu2024mobilellm} employs a multi-stage curriculum to master mobile app navigation, maintaining an internal visual state to reason about on-screen affordances and determine subsequent actions.

A significant trend is the development of unified vision-language-action (VLA) models that operate end-to-end. These agents use only visual observations and closed-loop reasoning to interact with GUIs. For example, ShowUI~\citep{lin2024showui} introduces a lightweight VLA model that processes screenshots and outputs UI actions through interleaved token modeling, achieving strong zero-shot accuracy. In a similar effort, UI-TARS~\citep{qin2025ui} functions as a pure vision agent, executing actions based solely on screenshots, while its System-2 reasoning module handles task decomposition. Aguvis~\citep{xu2024aguvis} further standardizes GUI interaction across platforms using only pixel inputs, training on a multimodal trajectory dataset and using self-generated reasoning steps to guide its decisions. These models exemplify closed-loop visual reasoning by observing the screen, interpreting its structure, and generating executable actions without textual UI representations.

A primary challenge for GUI agents is obtaining sufficient training data to cover diverse applications and user goals. To address this, several data acquisition strategies have been explored. Explorer~\citep{pahuja2025explorer} uses automated web exploration to generate a large dataset of task trajectories, significantly boosting web automation performance. TongUI~\citep{zhang2025tongui} converts online GUI tutorials into multimodal training data via automatic extraction. PC Agent-E~\citep{he2025efficient} demonstrates that few-shot expert demonstrations, augmented with synthetic variations, can enable agents to achieve strong generalization and outperform large-scale baselines.

\paragraph{\includegraphics[scale=0.03]{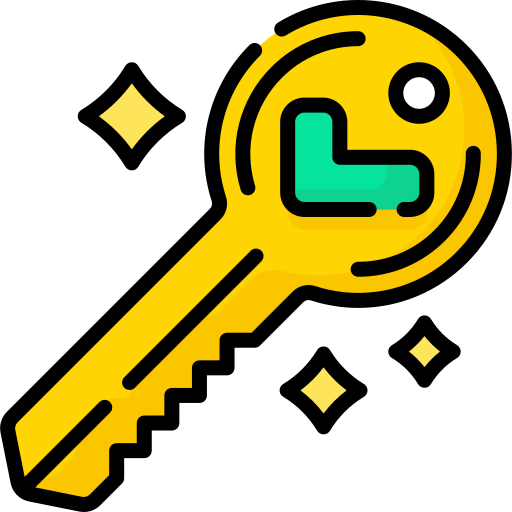} Takeaway.}
The application of ``Thinking with Images'' to GUIs marks a paradigm shift from reliance on structured data like HTML to direct, pixel-based reasoning. This enables agents to operate across diverse platforms such as web, desktop, and mobile environments where programmatic access is unavailable. The core advancement lies in creating a closed-loop cognitive process. Models observe the visual state of an interface, interpret its affordances, and generate the next action, often using an internal visual chain-of-thought to plan multi-step tasks. This iterative cycle of perception and action allows for real-time, multi-turn control. While the diversity of interfaces poses a significant data challenge, recent work demonstrates promising solutions through automated data generation from web exploration, conversion of online tutorials, and efficient learning from a few expert demonstrations. These developments are paving the way for more general and capable UI agents that interact with digital environments as humans do.

\subsection{Embodied AI and Robotics}
\label{ssec:app_embodied_ai}

To operate effectively in unstructured physical environments, embodied agents such as robots must reason with visual inputs. Recent advances in LMMs are enabling visual thinking for action planning, from sensor-grounded language understanding to internal scene simulation~\citep{Driess2023PaLME,Fan2025InterleaveVLAa,zhao2025cot,Zawalski2025Robotic,zhou2025anyprefer}. Pioneering work like PaLM-E~\citep{Driess2023PaLME} demonstrated that integrating language with visual and sensor inputs allows for positive knowledge transfer from web-scale pretraining to real-world robot manipulation. Subsequent models like Interleave-VLA~\citep{Fan2025InterleaveVLAa} empower robots to process interleaved visual-language prompts and generate executable action sequences in zero-shot settings, improving generalization to unseen objects. Similarly, VIMA~\citep{Jiang2023VIMA} advances general-purpose robot control by using multimodal prompts to support open-vocabulary manipulation, enabling robots to interact with novel objects through LMM-generated structured cues.

Beyond direct perception-to-action policies, many approaches incorporate a visual chain of thought to enhance planning. For example, CoT-VLA~\citep{zhao2025cot} improves temporal reasoning by prompting the model to predict intermediate visual subgoals before generating an action, which achieves substantial performance gains in manipulation tasks. Embodied ECoT~\citep{Zawalski2025Robotic} introduces a multi-step reasoning framework that explicitly considers plans, object features, and robot state before action prediction, significantly improving policy generalization. Pushing this concept further, VPRL~\citep{xu2025visualplanningletsthink} conducts planning entirely through sequences of generated images, akin to mental sketches. In spatial navigation tasks, VPRL outperformed language-based methods, suggesting that image-only reasoning may be more effective for certain spatial problems.

Another critical direction involves learning visual world models that simulate future states, allowing agents to reason through visual imagination. Genie~\citep{Bruce2024Genie} introduces a method to learn an interactive world model directly from internet videos, enabling frame-by-frame world generation in an unsupervised manner. For robotics, 3D-VLA~\citep{Zhen20243DVLA} builds a 3D goal-conditioned world model using diffusion, generating future goal states to facilitate visualized intent planning. UniSim~\citep{Yang2024Learning} learns from real-world interaction videos to simulate the visual outcomes of actions, enabling experience generation for policy learning. These works aim to develop a ``Mind's Eye'' for embodied agents, enabling them to imagine future outcomes and act on internal simulations.

\paragraph{\includegraphics[scale=0.03]{picture/key.png} Takeaway.}
In embodied AI, the ``Thinking with Images'' paradigm provides a crucial bridge between high-level reasoning and low-level physical action. Models are moving beyond simple perception-to-action mappings and adopting more deliberative processes. One key approach is the use of a visual chain of thought, where agents generate intermediate visual subgoals or reasoning steps to guide their planning. This enhances performance and interpretability in complex manipulation tasks. A more advanced frontier is the development of visual world models. These models empower agents with imagination, allowing them to simulate future visual states and predict the outcomes of their actions. This internal simulation capability is fundamental for long-horizon planning and robust decision-making in dynamic environments. Furthermore, some research indicates that for spatial tasks like navigation, reasoning with generated images may be more natural and efficient than using textual descriptions, pointing toward a future of ``visual-native'' robotic intelligence.

\subsection{AI for Science}
\label{ssec:app_ai_for_science}

LMMs are beginning to support scientific inquiry by interpreting complex visual data from experiments, microscope slides, and satellite imagery~\citep{yan2025position, zhu2025maps, tan2025chemmllm, burgess2025microvqa}. However, a consistent challenge emerges across these domains. While general-purpose models can describe visual scenes, they often fail to grasp the underlying physical dynamics and domain-specific principles crucial for rigorous scientific reasoning~\citep{SchulzeBuschoff2025Visual}. The ``Thinking with Images'' paradigm directly addresses this gap by empowering models with a deeper, more active form of visual cognition that is grounded in scientific principles.

This advanced capability allows models to move beyond simple description to active analysis. For example, in remote sensing, a model can think with images by invoking a specialized tool to precisely quantify deforestation, a functionality explored in systems like GeoChat which supports complex geospatial dialogues~\citep{Kuckreja2023GeoChat}. In fields like microscopy where experts can easily interpret cluttered scenes that challenge LMMs~\citep{Verma2024Human}, a model thinking with images could generate code to segment and highlight specific cell types. This creates a new, clarified visual representation that serves as a verifiable intermediate step in its analytical process.

The most profound application lies in using visual imagination for simulation. In physics, for instance, a model can learn to simulate the outcomes of physical events, such as predicting the stability of a block tower~\citep{Balazadeh2024Synthetic}. This transforms the model from a passive observer into an active simulator that can conduct ``virtual experiments''. While current models are still developing this expertise, they are laying the foundation for future scientific agents that do not just parse data, but actively reason and experiment with it visually to uncover new insights~\citep{bercovich2025llama}.

\paragraph{\includegraphics[scale=0.03]{picture/key.png} Takeaway.}
The application of ``Thinking with Images'' in science is in its nascent stages, but it already holds significant promise for automating data analysis and hypothesis generation. Current LMMs show foundational capabilities in interpreting diverse scientific imagery, including physics simulations, microscopy slides, and remote sensing data. However, a critical finding is that general-world knowledge is insufficient for specialized scientific domains. The performance of these models improves dramatically with domain-specific adaptation. Training on curated datasets, such as physics simulations or annotated satellite images, allows the models to learn the relevant scientific priors and visual patterns that general models miss. This points out a key insight: for AI to become a valuable partner in science, it must not only see but also understand the visual world through the lens of specific scientific disciplines. Future progress will likely depend on developing models that can integrate this domain knowledge more deeply, enabling them to perform more complex visual reasoning and even generate novel visual hypotheses.

\subsection{Healthcare and Medicine}
\label{ssec:app_healthcare}

In healthcare, LMMs are evolving from passive image interpreters to active partners in clinical decision-making~\citep{li2023llava, moor2023med, chen2024towards, xia2024cares, xia2024rule, seyfioglu2024quilt, xia2025mmed, zhu2025mmedpo, nath2025vila, lai2025med, lin2025healthgpt, dai2025qoq}. The ``Thinking with Images'' paradigm is central to this evolution, enabling models to generate dynamic and explainable multimodal medical reasoning rather than static predictions. Early systems demonstrated this by combining vision models with LLMs to produce detailed diagnostic reports for chest X-rays, improving both accuracy and patient communication~\citep{Wang2024Interactive}.

A key challenge in this high-stakes domain is ensuring reliability. ``Thinking with Images'' offers powerful new strategies to enhance trustworthiness. For instance, models can be guided by expert knowledge, such as using anatomical ontologies to direct a multi-step visual search across a medical scan~\citep{li2025aor, Guo2025Prompting}. This structured approach mitigates hallucination and mimics clinical workflows. For highly specialized fields like digital pathology, models are learning to programmatically construct visual hierarchies of cells and tissues, enabling precise analysis of gigapixel slides~\citep{Dai2025Pathologyvlm}. These methods transform the model from a black box into a transparent reasoner. At the forefront of this domain, applications are beginning to model complex cognitive processes, either by simulating clinical consultations through multi-agent collaboration or by intrinsically generating visual hypotheses. Frameworks like MMedAgent-RL simulate a consultation between a generalist and specialist AI, using dialogue to refine a diagnosis~\citep{xia2025mmedagent}. This represents a collaborative form of visual thought. Furthermore, generalist models like HealthGPT are beginning to natively synthesize new visual content, such as generating a healthy organ scan for comparison, as part of a unified reasoning process~\citep{lin2025healthgpt}. Efforts to evaluate these advanced capabilities are also emerging, with benchmarks like MedEBench assessing the fidelity of text-guided clinical image editing~\citep{liu2025medebenchrevisitingtextinstructedimage}. 

\paragraph{\includegraphics[scale=0.03]{picture/key.png} Takeaway.}
In the medical domain, ``Thinking with Images'' is transforming diagnostic AI from a predictive tool into a collaborative reasoning partner. This paradigm grounds abstract medical analysis in concrete visual evidence, enabling models to generate interpretable reports, follow anatomical search paths, and even simulate clinical consultations. The critical insight from this high-stakes field is that the primary frontier is not just capability, but verifiability and safety. Therefore, the most promising advancements involve constraining and guiding the model's visual thought process with expert knowledge and structured workflows. The ultimate objective is to develop AI systems that reason visually with clinical rigor, functioning as dependable assistants whose diagnostic conclusions are not only accurate but also transparent and auditable.

\subsection{Education and Training}
\label{ssec:app_education}

LMMs are transforming education by enabling AI tutors that ``Thinking with Images'', combining visual reasoning with natural language to create interactive learning experiences. A representative system is Interactive Sketchpad~\citep{chen2025interactive}, which assists students in solving STEM problems by pairing step-by-step textual explanations with dynamic visual illustrations. For example, it can guide a student through a geometry proof by drawing auxiliary lines on a diagram while narrating each logical step. This approach, which integrates code-based diagram generation to function as a virtual whiteboard, improves student engagement and accuracy. The underlying capability for such systems is driven by benchmarks like MathVista~\citep{lu2023mathvista}, which push for advances in visual mathematical reasoning, and specialized models like DiagramGPT~\citep{Zala2024DiagrammerGPT}, which focuses on generating open-domain diagrams through a two-stage planning and rendering process.

Beyond K-12 education, LMMs are advancing professional training through visual feedback systems. VidAAS~\citep{Lee2024see} uses GPT-4V to analyze classroom recordings and provide detailed assessments of teaching performance. It interprets non-verbal cues like body language and comments on spatial arrangements to deliver context-aware feedback that helps educators improve. Similar applications are emerging in medical training, where virtual patients could be driven by LMMs that interpret a trainee's gestures or visual cues, simulating clinical scenarios with realism~\citep{Singhal2023Large}. These examples demonstrate a shift from static instructional content to dynamic, visually-grounded educational agents delivering personalized, interactive learning experiences.

\paragraph{\includegraphics[scale=0.03]{picture/key.png} Takeaway.}
In education, the ``Thinking with Images'' paradigm enables new AI tutors that do more than provide textual answers; they can visually demonstrate concepts. By generating diagrams, highlighting key features, or sketching solutions in real-time, these systems function as interactive whiteboards, making abstract subjects like mathematics more intuitive. This dynamic visual dialogue adapts to a learner's needs, offering a more engaging and personalized educational experience. The application extends to professional training, where models provide detailed feedback by analyzing visual data from real-world scenarios like classroom videos. This capability for contextual visual analysis allows feedback previously only possible through human observation. These advances signal the emergence of visually-grounded educational agents blending image and language understanding to deliver more effective, interactive, and human-like learning support.

\section{Future Directions}

Although ``Thinking with Images'' marks a paradigm shift in artificial intelligence, we are still at the early stages of realizing its full potential. Looking forward, this section explores the pivotal research avenues that will propel the field into its next era, focusing on the innovations required to unlock more efficient, secure, and profoundly capable forms of visual cognition.

\label{sec:challenges}
\subsection{Computational and Cognitive Efficiency}
\label{ssec:Computational}

A major challenge for ``Thinking with Images'' is its high computational cost~\citep{ning2024inf, zhou2024survey}. Current methods often rely on a sequence of visual steps, such as calling an external tool or generating an intermediate image. Each step adds significant delay and requires a lot of computation, making these models too slow for real-time applications~\citep{liu2024mobilellm}. Future research should find ways to achieve powerful visual reasoning without costly execution. This could involve teaching models to compress a long reasoning process into fewer steps, or to predict the final outcome without needing to render every single intermediate image~\citep{lee2025well}.

A deeper challenge is improving cognitive efficiency, which is about using the right amount of effort for a given task~\citep{qu2025survey, huang2025adactrladaptivecontrollablereasoning}. A truly smart system should not use its most powerful and expensive thinking abilities for every simple problem~\citep{wang2025think}. The next generation of models must learn to decide when a quick textual answer is enough, and when a more careful visual analysis is actually needed. This is like a human deciding whether to solve a math problem in their head or to pull out a piece of paper and a pencil. Future work could train models with reward systems that penalize unnecessary visual steps, encouraging them to be more resourceful. The goal is to move from models that follow a fixed procedure to models that can flexibly switch between fast, simple thinking and slow, deliberate visual reasoning.

Achieving this dual efficiency will also require new model architectures~\citep{deng2025bagel, chen2025blip3}. Instead of calling slow external tools, future models could have small, built-in modules designed for common visual tasks, like searching for an object or comparing two elements. These specialized modules could work directly on the model's internal data representations, or latent space, avoiding the slow process of creating and analyzing full pixels. The objective is to develop models that can think with images while also demonstrating a sense of purpose and economy, making advanced visual reasoning a scalable and accessible technology.

\subsection{Safe and Trustworthy Visual Cognition}
\label{ssec:safe}

The ability to ``Think with Images'' creates new safety and ethical challenges~\citep{andriushchenko2024jailbreaking, qu2024alleviating}. The most direct danger is the potential for misuse. Because these models can generate a series of images that seem to form a logical argument, a malicious user could guide a model to create a visual story that supports a false claim~\citep{nguyen2022deep}. This goes beyond a single deepfake. It enables the automated creation of entire disinformation campaigns, where fabricated visual evidence is presented step-by-step to make a lie seem true. The challenge, therefore, is not just blocking harmful images, but ensuring the entire thinking process cannot be hijacked to build a misleading narrative~\citep{bai2022constitutional}.

The security of these models is also a major concern due to their complex structure. The points of attack are no longer just the initial inputs, but the entire reasoning loop~\citep{liu2024survey, liudz2024survey, liu2024pandora}. An attacker could poison the external tools a model relies on, corrupting its analysis from the start. A more subtle attack could involve designing an input that tricks the model into making a bad choice, such as focusing on a distracting detail or selecting the wrong tool for the job. In this way, the threat moves from simply causing a wrong output to corrupting the very process of thinking itself. This requires new security measures to protect a model's decisions, tools, and internal imagination.

Finally, the thinking process of these models is often a "black box", which makes it hard to trust them. Hidden biases can affect not just the final output, but also the internal reasoning steps~\citep{howard2024uncovering}. For example, a model might learn to be more critical or generate stereotypical images only when reasoning about certain groups of people. This could create a hidden, self-reinforcing cycle of prejudice that is very difficult to spot. Making these systems trustworthy requires more than just explaining a result~\citep{aflalo2022vl}. We need new methods to see and review the model's internal conversation of tool use and image generation. Without this transparency into how a model thinks with images, we cannot reliably find and fix internal biases, which prevents us from building models whose reasoning is not only powerful but also provably fair~\citep{chang2024survey}.

\subsection{Novel Benchmarks and Evaluation Methodologies}
\label{ssec:novel}
A key problem for the ``Thinking with Images'' paradigm is that current benchmarks are inadequate~\citep{li2024survey}. Most datasets only check the final answer and do not examine the reasoning process. This creates a loophole: a model can get the right answer for the wrong reasons, for example by using shortcuts instead of genuinely thinking with the image~\citep{chen2024we}. To address this, future work should focus on two main areas: creating more complex scenarios that require thinking with images, and developing new ways to evaluate the thinking process itself.

The first area of future work is to build new benchmarks with tasks that cannot be solved without actively thinking with images. One important direction is to design interactive and multi-step problems. For example, a task could require a model to solve a virtual puzzle, where each action changes the visual scene and reveals the next clue. Another promising direction is to create a greater variety of meaningful and challenging tasks that require visual construction. This means the model must generate a new visual element to find the solution, such as drawing an auxiliary line on a geometry diagram or sketching a simple machine to solve a physics problem~\citep{wu2024star}. In these scenarios, simply looking at the image once is not enough; success depends on a continuous process of visual interaction and creation.

The second area of future work is to develop new methods for evaluating the thinking process, not just the final answer~\citep{song2025prmbench}. Instead of only using accuracy, future metrics must assess the quality of the ``Thinking with Images'' chain. For example, we need evaluation methods that can check if each step of a model's thought is grounded in the visual evidence, penalizing any steps that are made up. We also need to measure the logical coherence and necessity of the process. A practical way to test necessity is to remove an intermediate visual step generated by the model and see if it can still arrive at the correct solution. Establishing these process-oriented evaluation frameworks is a foundational step toward building more transparent and trustworthy multimodal AI systems.

\subsection{Thinking with Audio, Video, and the World}
\label{ssec:thinking}
While this survey has focused on static images, the paradigm’s critical extension is its evolution toward dynamic modalities like audio and video, which demand cognition that is not just spatial but temporally coherent. This gives rise to two parallel frontiers: ``Thinking with Audio'' and ``Thinking with Video''. In ``Thinking with Audio'', a model might reason about the emotional prosody of speech or simulate a soundscape to predict an event~\citep{latif2023sparks}. Likewise, ``Thinking with Video'' moves from static scenes to dynamic events, demanding a profound shift in cognitive capabilities. It requires not only passive viewing but also the autonomy to actively navigate the flow of time by deciding when to replay, fast-forward, or focus on critical moments to build a coherent understanding~\citep{maaz2023video}. This active navigation of the timeline extends beyond observed events, requiring the model to reason about the unseen through internal simulation. It might infer causality by constructing the unobserved events that connect disparate moments, or anticipate futures by generating plausible subsequent frames to predict the outcome of an action~\citep{wu2024next, chen2025counterbench, tong2025mj, hafner2023mastering}. The chain of thought thus evolves from a static script into a dynamic mental model that unfolds with the narrative, transforming the model from a passive processor of frames into an active cognitive observer of a temporal world.

The ultimate frontier of this research results in ``Thinking with the world'', where the paradigm shifts from detached observation to embodied agency \citep{brohan2023rt}. Here, reasoning becomes an integral component of a continuous perception-action loop. The model’s thoughts directly inform its physical actions, and the sensory consequences provide immediate feedback that grounds its understanding \citep{fu2024mobile}. Its generative power evolves into a mechanism for mental rehearsal, allowing the agent to simulate action sequences before committing to a physical choice. This process grounds abstract concepts in tangible reality, defining an object not by its appearance but by its physical affordances in a planned task \citep{liu2024ok}.

The paradigm’s most profound shift is making the reasoning process itself a manipulable object. Future progress depends on enriching this new ``language'' of thought. Achieving this synthesis promises truly autonomous agents that can compose their analytical steps into complex and reliable cognitive programs, allowing them to purposefully navigate and shape their environment.

\subsection{Open Questions and Outlook}

\label{sec:open_discussion}

The emergence of ``Thinking with Images'' opens profound questions about the nature of artificial cognition and its ultimate architectural form. To effectively guide future research, we first delineate the relationship between this internal cognitive paradigm and the external framework of generalist agents. We then propose a blueprint for a unified visual thinker, synthesizing the stages discussed in this survey into a cohesive, forward-looking vision.

\subsubsection{Thinking with Images \textit{v.s.} Agentic Frameworks}
\vspace{-5pt}
The ``Thinking with Images'' paradigm defines an internal cognitive process focused on \textit{how} a model reasons, whereas an agent framework provides an external execution cycle focused on \textit{how} a system acts. One is a mechanism for thought; the other, an architecture for action.
\vspace{-5pt}
\paragraph{Deliberation \textit{v.s.} Execution.}
The core distinction between the ``Thinking with Images'' paradigm and an agent framework is their fundamental objectives. The ``Thinking with Images'' paradigm is concerned with the fidelity and depth of the reasoning process itself. Its primary goal is to enhance understanding by generating and manipulating visual information to explore a problem space, verify hypotheses, and uncover insights that are difficult to articulate through language alone. An agent framework, conversely, is oriented toward external task completion. Its objective is to successfully execute a mission within an environment, where success is measured by the final outcome. The former prioritizes the quality of deliberation, while the latter prioritizes the efficacy of execution.
\vspace{-5pt}
\paragraph{Workspace \textit{v.s.} Perception.}
The distinct objectives of the ``Thinking with Images'' paradigm and an agent framework also lead to a fundamentally different role for visual information. For the ``Thinking with Images'' paradigm, vision serves as a dynamic and manipulable cognitive workspace. It functions as a mental sketchpad where ideas can be visually instantiated, modified, and examined as part of an unfolding reasoning sequence. For an agent, vision is primarily the medium of environmental perception. It provides a snapshot of the external world state, which must be interpreted to inform the agent’s next decision. In one context, the image is an internal and dynamic tool for thought; in the other, it is an external stimulus for a single action.

\paragraph{Artifacts \textit{v.s.} Instructions.}
The distinction between the ``Thinking with Images'' paradigm and an agent framework further manifests in the nature of their intermediate steps. The intermediate steps in a ``Thinking with Images'' process are cognitive artifacts. They are internal mental constructs, such as an imagined future scene, an annotated diagram to clarify spatial logic, or a digitally enhanced region of an image to reveal fine-grained details. These steps are not meant to act upon the world but to refine the model's internal understanding. The intermediate steps for an agent, however, are typically executable instructions. They are commands intended to have a direct effect, such as API calls, code execution, or signals sent to physical actuators.
\vspace{-6pt}
\paragraph{Engine \textit{v.s.} Machine.}
Rather than being mutually exclusive, the ``Thinking with Images'' paradigm and an agent framework share a interconnected and complementary relationship. The most effective way to conceptualize this is to view the ``Thinking with Images'' paradigm as the cognitive engine within the agent's machine. An agent provides the essential architecture for worldly interaction, equipped with perception sensors, action effectors, and memory systems. The ``Thinking with Images'' paradigm provides the advanced reasoning power that makes the agent’s actions intelligent rather than merely reactive. This synergy becomes tangible when we consider how the stages outlined in this survey empower an agent's capabilities. An agent leverages Stage 1 by learning to orchestrate external visual tools. It evolves by using Stage 2 to programmatically analyze visual data. Ultimately, a truly advanced agent embodies Stage 3, using intrinsic imagination to simulate the consequences of potential actions. Therefore, the research presented in this survey is not an alternative to agent-based AI but is instead a foundational step toward building autonomous systems that possess a deeper, more human-like capacity for visual understanding and intelligent action.

\begin{figure*}[t]
    \centering
    \includegraphics[width=0.9\textwidth]{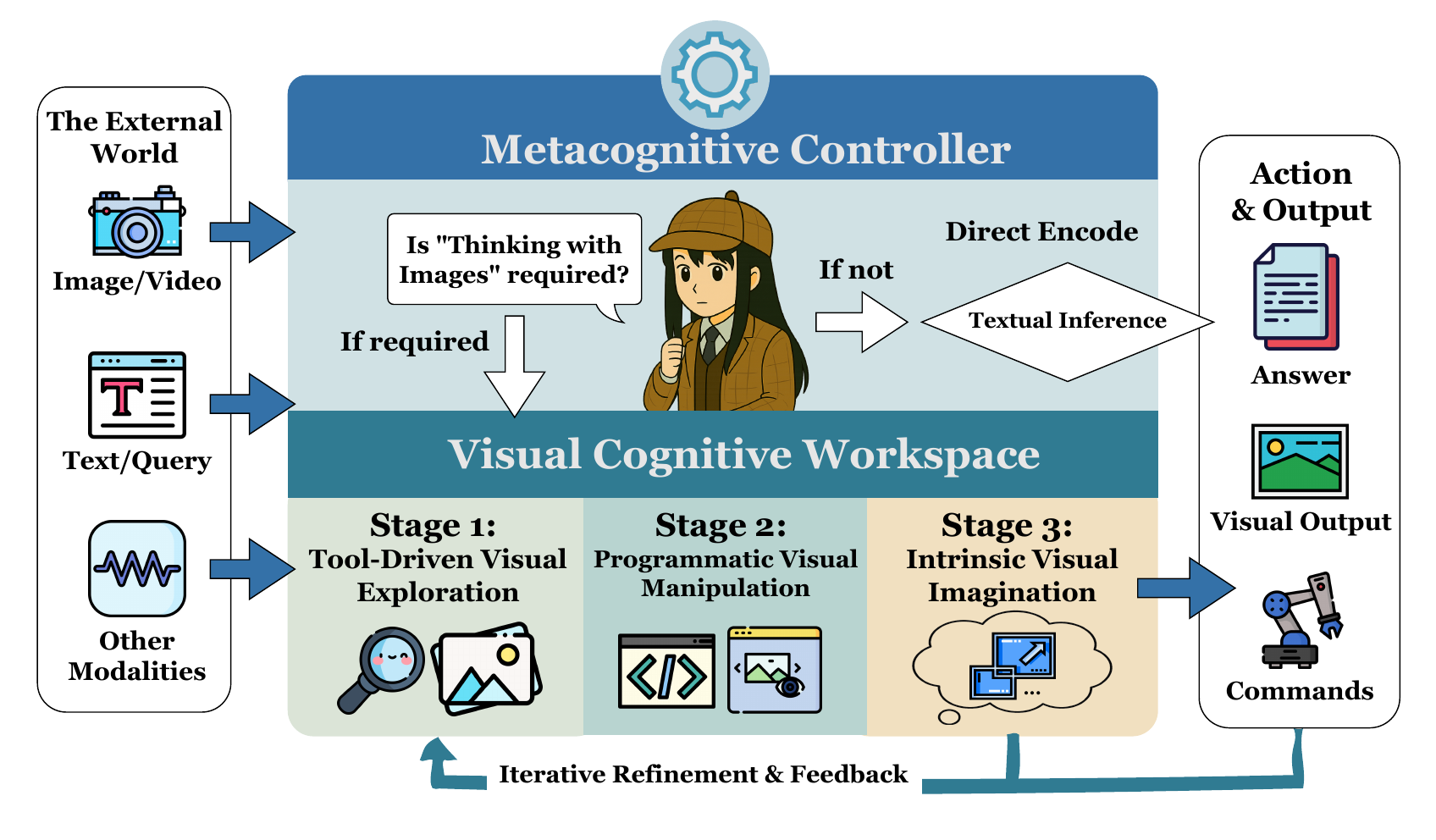}
    \vspace{-6pt}
    \caption{A conceptual blueprint for a Unified Visual Thinker. The architecture features a Metacognitive Controller that directs tasks to either an immediate inference pathway or a multi-stage Visual Cognitive Workspace, enabling flexible and efficient multimodal reasoning.}
    \label{fig:thinker}
    \vspace{-6pt}
\end{figure*}

\subsubsection{Blueprint for a Unified Visual Thinker}
\label{sssec:blueprint}

In the final section, we propose a conceptual blueprint for a unified visual thinker, as illustrated in Figure~\ref{fig:thinker}. This architecture synthesizes the core ideas of our survey into a single, cohesive framework, outlining a path toward a more capable and efficient form of multimodal AI. The central principle is not to create a model that always relies on its most complex reasoning, but a dynamic system that intelligently selects the right cognitive tool for the task.

\paragraph{Metacognitive Controller.}
The Metacognitive Controller serves as the system's decision-making core. It receives inputs from the external world and its primary role is to assess the task's complexity. For straightforward problems, it chooses a simple, direct path: the model generates an answer based on a quick, immediate interpretation of the visual data. This approach avoids unnecessary computational cost for tasks that do not require it. For more complex problems that demand deeper analysis, the controller activates the Visual Cognitive Workspace, signaling the need for a more deliberate, visually-grounded thought process.

\paragraph{Visual Cognitive Workspace.}
The Visual Cognitive Workspace is the active environment where this deeper reasoning happens. Guided by the controller, it can dynamically operate across the three stages of cognitive autonomy: leveraging Stage 1 to explore with tools, Stage 2 to create programmatic analyses, or Stage 3 to imagine and simulate outcomes. A key feature of this workspace is the feedback loop, which allows the system to review and build upon its own generated thoughts. However, this process is flexible; the system is not required to complete a full cycle and can produce a final output at any point if it determines a satisfactory solution has been reached.
\vspace{-5pt}
\paragraph{Action \& Output Interface.}
The Action \& Output interface is the final component, responsible for translating the system's internal thoughts into concrete actions. This is where the model's reasoning connects back to the external world. The outputs can take various forms depending on the task: a textual answer for a question, a new visual output like an edited image or a diagram, or a sequence of executable commands that could guide a robotic agent.

\vspace{-5pt}
\paragraph{A Unified Vision.}
This integrated architecture provides the foundation for an agent that embodies the ultimate promise of this survey: an AI that does not merely think about images, but truly thinks with them. The synergy between the Metacognitive Controller and the multi-stage Visual Cognitive Workspace enables a crucial cognitive flexibility, allowing the model to dynamically transition between tool-driven exploration, programmatic creation, and deep, internal simulation. This approach moves AI beyond text-centric reasoning, adding a dynamic visual component to its ``language of thought''. The key to a true visual thinker, then, is not just the ability to imagine, but the meta-reasoning needed to choose the most effective and efficient path for a given task. This blueprint provides a clear direction for building more powerful and genuinely multimodal AI systems.

\section{Conclusion}
\label{sec:conclusion}

This survey has charted the evolution of multimodal reasoning towards a new paradigm: ``Thinking with Images''. We have systematically organized this emerging field into a three-stage framework, detailing the progression from models that leverage external tools, to those that perform programmatic manipulation, and finally to systems also capable of intrinsic imagination. Our analysis reveals a fundamental shift in the role of vision from a static input to a dynamic cognitive workspace. This transformation enables deeper perceptual analysis and more robust physical simulation, laying the groundwork for more intuitive and powerful AI systems.

The journey towards true visual cognition is still in its early stages. Significant challenges in efficiency, robustness, and generalization must still be overcome. The path forward requires developing more well-developed world models, exploring abstract visual representations beyond pixels, and building unified frameworks for human-AI collaboration. Ultimately, the research presented here marks a foundational step toward creating AI agents that not only see the world, but reason within it as well, equipped with a mind's eye that is integral to their intelligence.
%%%%%%%%%%%%%%%%%%%
%%%%%%%%%%%%%%%%%%%%%%%%%%%%%%%%%%%%%%%%%
\section*{Contributions}

The contributions of all authors are listed as follows: Zhaochen Su drafted the abstract, introduction (\S\ref{sec:intro}), foundations (\S\ref{sec:foundations}), and the introductory sections for each stage (\S\ref{sec:stage1-formulation}, \S\ref{sec:stage2-leap}-\ref{sec:stage2-formulation}, \S\ref{sec:stage3-leap}-\ref{sec:stage3-formulation}). For the core methods, authorship was divided by approach: Zhenhua Liu drafted the prompting sections (\S\ref{sec:stage1-Approaches-Prompt-Based}, \S\ref{sec:stage2-Approaches-Prompt-Based}), Peng Xia the SFT sections (\S\ref{sec:stage1-Approaches-SFT-Based}, \S\ref{sec:stage2-Approaches-SFT-Based}, \S\ref{sec:stage3-Approaches-SFT-Based}), and Yan Ma the RL sections (\S\ref{sec:stage1-Approaches-RL-Based}, \S\ref{sec:stage2-Approaches-RL-Based}, \S\ref{sec:stage3-Approaches-RL-Based}). Additionally, they authored key framing sections for their respective stages: Zhenhua for Stage 1 (\S\ref{sec:stage1-categories}, \S\ref{sec:stage1-conclusion}), Peng for Stage 2 (\S\ref{sec:stage2-categories}, \S\ref{sec:stage2-conclusion}), and Yan for Stage 3 (\S\ref{sec:stage3-categories}, \S\ref{sec:stage3-conclusion}). Moreover, Hangyu Guo drafted \S\ref{sec:evaluation}, and Jiaqi Liu drafted \S\ref{sec:application}. Zhaochen Su and Xiaoye Qu drafted \S\ref{sec:challenges}.

For visual elements, Zhaochen Su created the abstract figure and  Figures~\ref{fig:compare_paradigms}, \ref{fig:evolve}, and \ref{fig:thinker}. Figure~\ref{fig:think-with-image-survey} was contributed by Peng Xia, Zhenhua Liu, and Kaide Zeng, while Figure~\ref{fig:benchmark_examples} was contributed by Hangyu Guo. Yanshu Li provided visual assets and initial drafts. All tables (Tables~\ref{tab:stage1_methods}, \ref{tab:stage2_methods}, \ref{tab:stage3_methods}) were a collaborative effort by Zhaochen, Zhenhua, Peng, Yan, and Kaide.

Junxian He and Yi R. (May) Fung  supervised the project, providing invaluable guidance on its overall direction, structure, and refinement. Zhengyuan Yang, Linjie Li, Yu Cheng, and Heng Ji provide insightful feedback and critical suggestions on the manuscript.
\clearpage

\bibliography{neurips_2025}
\bibliographystyle{unsrtnat}

\end{document}